\pdfoutput=1

\documentclass[10pt,twocolumn,letterpaper]{article}

\usepackage[pagenumbers]{cvpr} %

\usepackage{colortbl}
\definecolor{rankbase}{HTML}{9BBECF}
\colorlet{rankfirst}{rankbase}
\colorlet{ranksecond}{rankbase!55}
\colorlet{rankthird}{rankbase!25}
\definecolor{vbgray}{gray}{0.93}  %
\AtEndPreamble{
    \crefname{figure}{Fig.}{Figs.}
    \Crefname{figure}{Fig.}{Figs.}
    \crefname{table}{Tab.}{Tabs.}
    \Crefname{table}{Tab.}{Tabs.}
}

\definecolor{cvprblue}{rgb}{0.21,0.49,0.74}
\usepackage[pagebackref,breaklinks,colorlinks,allcolors=cvprblue]{hyperref}

\title{{\ourmethod}: Anything Can Represent Camera for Video Generation}

\author{Weijie Lyu$^{1}$ \quad Zhixin Shu$^{2}$ \quad Yuntian Ke$^{3}$ \quad Ming-Hsuan Yang$^{1}$\\[2mm]
$^{1}$University of California, Merced \quad $^{2}$Adobe Research \quad $^{3}$University of Pennsylvania
}

\newcommand{\ourmethod}{\textit{AnyCam}}
\newcommand{\ourmethodar}{\textit{AnyCam-AR}}

\begin{document}
\captionsetup{hypcap=false}
\twocolumn[{%
\renewcommand\twocolumn[1][]{#1}%
\maketitle
\begin{center}
\vspace{-4mm}
\centering
\includegraphics[width=0.98\linewidth]{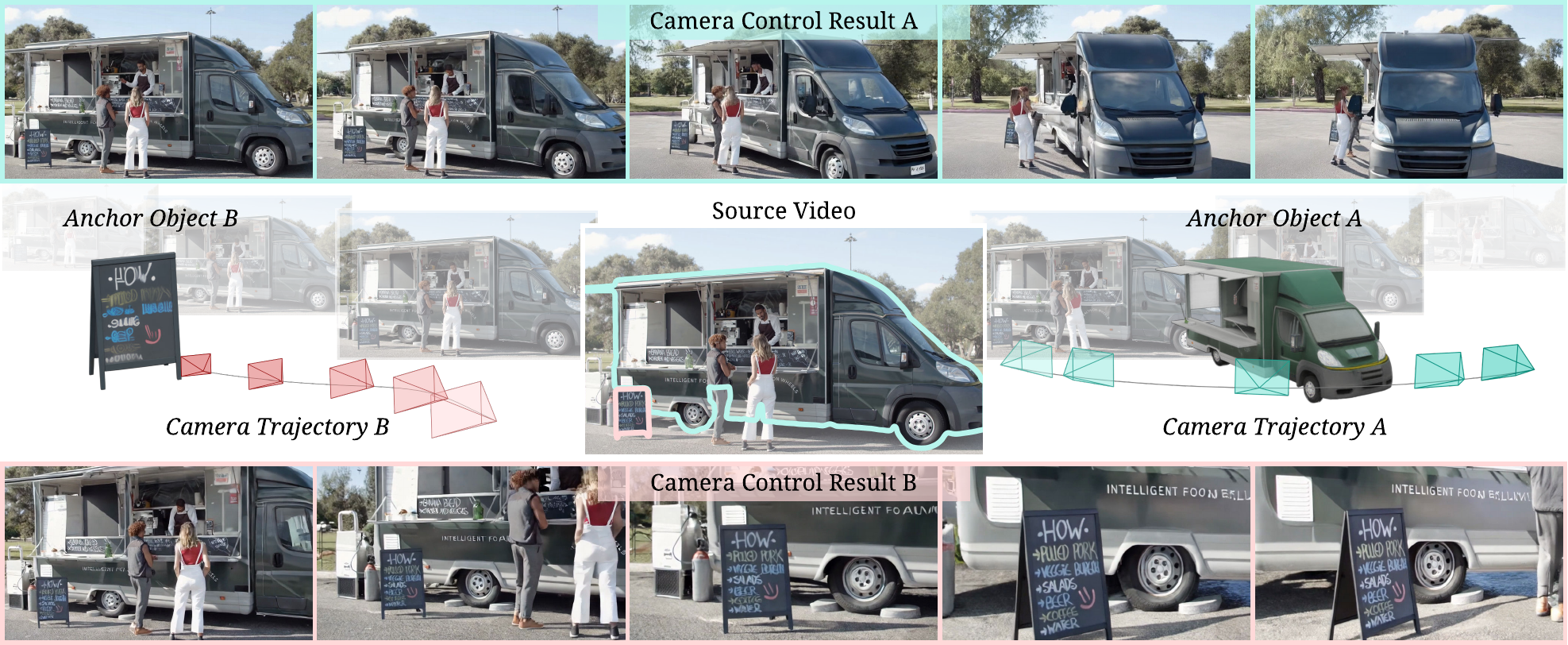}
\vspace{-2mm}
\captionof{figure}{{\ourmethod} moves the camera of a video relative to any object in the scene. Given a source video (middle), the user picks an anchor object, here the food truck (A) or the chalkboard sign (B), and a camera trajectory around it. We reconstruct the anchor in 3D and render it along the trajectory, and this rendering conditions the camera-controlled video generation. Orbiting the truck turns the view to its front (top row), and moving toward the sign brings its menu into view (bottom row), while the people and their motion follow the source video.}
\label{fig:teaser}
\end{center}
}]
\begin{abstract}
Camera-controlled video generation re-renders a video along a target camera trajectory, yet moving the camera around an object, the most natural way to change the viewpoint of an object-centric scene, remains difficult.
Existing methods condition either on camera parameters, which are scale-ambiguous in monocular video, or on explicit geometry such as warped point clouds or depth, which breaks down under large viewpoint changes.
We present {\ourmethod}, a method that represents the camera with an anchor object in the scene: we reconstruct it in 3D and render it along the target trajectory, a camera signal that is scale-aware, view-complete and easy to specify relative to the object.
A video diffusion model transfers the content and dynamics of the source video to the new viewpoint.
Its H-Concat layout aligns source and target frames in time, so the model can be distilled into {\ourmethodar}, a streaming few-step generator for on-the-fly camera control.
Both models are trained on paired videos with pixel-aligned anchor renderings from multi-camera videos and static-scene captures.
Experiments on diverse datasets and trajectories show that {\ourmethod} performs favorably against camera-controlled video generation and 3D reconstruction methods in camera control and visual quality. Project page: \url{https://weijielyu.github.io/AnyCam}.
\end{abstract}

\vspace{-3mm}
\section{Introduction}
\vspace{-1mm}
\label{sec:intro}
World modeling has been a central topic in computer vision for decades, with diverse applications in simulation, AR/VR, and robotics.
One of its key tasks is novel view synthesis (NVS), where the goal is to predict renderings from viewpoints that are not in the input data.
NVS has long been addressed with neural rendering methods, such as NeRF~\cite{nerf}, Gaussian Splatting~\cite{gaussian_splatting} and feed-forward reconstruction models~\cite{lrm, pixelsplat, gslrm, facelift, instainpaint, edit3r}.
With the development of video generation, seminal works~\cite{gcd, recammaster, trajectorycrafter} have shown that video models with camera control ability can also be used for scene-level NVS.

With neural rendering methods, NVS is studied both for single objects and for whole scenes.
Camera-controlled video generation, however, has focused on scene-level trajectories such as pan, tilt, zoom and translation, while more challenging trajectories that travel around an object remain hard to accomplish.
We refer to this task as object-centric camera control, and attribute the difficulty mainly to the camera representation.

\begin{figure}[t]
  \centering
  \includegraphics[width=0.98\linewidth]{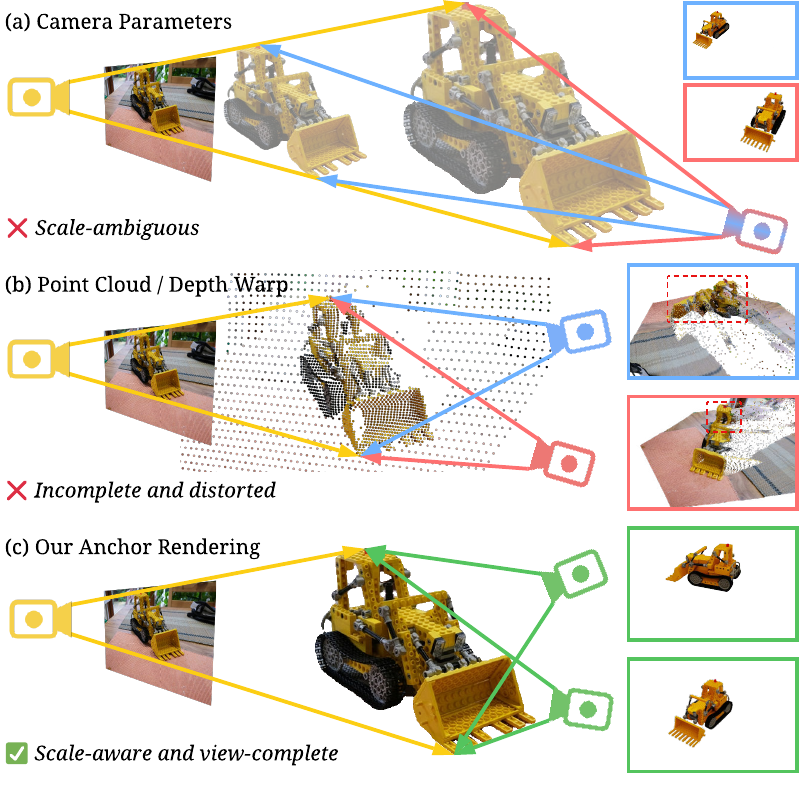}
  \vspace{-4mm}
  \caption{\textbf{Camera representations.}
  Yellow represents the input camera.
  (a) Camera parameters are scale-ambiguous: a small object near the input camera and a large object farther away give the same input image, but the same target camera parameters (blue and red camera) see them at a different size, position and angle (blue and red boxes), so the target view is not determined.
  (b) A point cloud or depth warp of the input image only covers the surface seen from the input camera.
  Under a large viewpoint change, views are often heavily distorted (red box). The points in the target view sometimes come from the inside of the object or from the back of the image (blue box). The renderings are wrong and mislead the subsequent video generation.
  (c) Our anchor rendering uses the complete reconstructed object, so target cameras at any angle (green) see the whole anchor with the correct size and pose.}
  \vspace{-4mm}
  \label{fig:cam_rep}
\end{figure}

Most of the existing methods represent the camera in two ways, and both struggle when the camera moves around an object (\cref{fig:cam_rep}).
Implicit representations~\cite{cameractrl, recammaster} inject the camera parameters into the network.
As analyzed in FaceCam~\cite{facecam}, a monocular video determines the scene only up to a global scale, so the camera parameters are scale-ambiguous: the same parameters are consistent with infinitely many scales and locations, and the generated motion is not determined.
Explicit representations~\cite{trajectorycrafter, track2view, geostream} instead render the camera motion as point clouds, depth or 2D tracks estimated from the input.
They inherit the errors of the estimated geometry. More fundamentally, they are not view-complete: they are empty wherever the input does not observe the scene, \eg, the back of an object, which is often the target of object-centric trajectories.

To address object-centric applications, we introduce {\ourmethod}, a camera-controlled video generation method that uses an object in the scene as the anchor that represents the camera.
Given a source video or an image, we select an anchor object in the scene, reconstruct it with SAM 3D Objects~\cite{sam3d}, and render it along the target trajectory to represent the viewpoint changes.
This \textit{anchor rendering} is (1) scale-aware, since the size of the anchor gives a consistent scale; (2) view-complete, since the reconstruction is a complete object; and (3) easy to specify relative to the object, \eg a $90^\circ$ orbit around the anchor.

A video-to-video model~\cite{wan} conditioned on the source video and the anchor rendering learns to read the target camera from the rendering and to take the geometry and appearance of the scene from the source video.
To train it, we construct source and target pairs from synchronized multi-camera videos and real static-scene captures, in which the anchor rendering is pixel-aligned with the target video.
The anchor is chosen automatically by a vision-language model~\cite{qwen3vl} when building the training data, and at inference the user can pick any suitable object.

Our model concatenates the target and the source along the image height, which we refer to as \textit{H-Concat conditioning}.
The frames of both videos at the same time step share one temporal position, which enables distilling {\ourmethod} into {\ourmethodar}, a few-step autoregressive video-to-video model.
Since every chunk of the output depends only on the matching chunk of the source, {\ourmethodar} accepts a streaming video as input, and the user can choose an anchor and steer the camera around it interactively.

We evaluate our method on diverse datasets, including monocular videos~\cite{davis_2016,davis_2017,re10k} and multi-view captures~\cite{mipnerf360, co3d} with ground truth, with both conventional camera trajectories and object-centric ones of diverse lengths.
The contributions of this work are:
\begin{itemize}
    \item We propose to represent the camera with the rendering of any object in the scene, a camera condition that is scale-aware, view-complete and object-relative, which enables object-centric camera control.
    \item We build a data pipeline that reconstructs the anchor, aligns it to the dataset cameras and produces source and target video pairs with pixel-aligned anchor renderings.%
    \item We design a time-synchronized H-Concat conditioning that aligns the source and the target in time, and use it to build {\ourmethodar}, a few-step streaming autoregressive model for interactive camera control.
    \item Experiments on diverse benchmarks show that {\ourmethod} performs favorably against camera-controlled video generation methods and 3D reconstruction methods in camera control and visual quality.
\end{itemize}

\vspace{-2mm}
\section{Related Work}
\vspace{-2mm}
\label{sec:related_work}
\noindent\textbf{Camera-controlled video generation.}
Camera-controlled video generation aims to synthesize videos following desired camera movements.
MotionCtrl~\cite{motionctrl} injects camera extrinsics. CameraCtrl~\cite{cameractrl}, CameraCtrl II~\cite{cameractrlII} and Wan2.2-Fun-Control-Camera~\cite{wan} encode the camera as Plücker rays.
ReCamMaster~\cite{recammaster} concatenates source and target tokens along the frame dimension under camera conditioning, and ReDirector~\cite{redirector} introduces a Rotary Camera Encoding derived from source and target rays.
Another line of work expresses the camera through images rendered or warped into the target view.
ViewCrafter~\cite{viewcrafter}, TrajectoryCrafter~\cite{trajectorycrafter} and GEN3C~\cite{gen3c} render point clouds estimated from the source and use them, together with the source video, as conditions.
DepthDirector~\cite{depthdirector} separates warped-depth guidance from source-video content, and ReCapture~\cite{recapture} renders a coarse target-view video and refines it with masked video fine-tuning.
FaceCam~\cite{facecam} renders a generic head along the target trajectory and extracts scale-aware facial-landmark maps, which gives accurate control for portrait videos.
{\ourmethod} also conditions on a rendered visual signal, but renders a 3D reconstruction of an object in the scene rather than warping the source frames.

\vspace{1mm}
\noindent\textbf{Streaming video generation and world models.}
Autoregressive video models generate successive frames or chunks from preceding context.
Diffusion Forcing~\cite{diffusion_forcing} trains causal denoising with independent per-token noise levels, CausVid~\cite{causvid} distills a bidirectional teacher into a few-step causal generator, and Self Forcing~\cite{self_forcing} trains on autoregressively generated histories to close the gap between training and inference.
Interactive world models condition generation on actions and retain a memory of past observations~\cite{nwm,worldmem,vmem}; HY-WorldPlay~\cite{worldplay} and InSpatio-World~\cite{inspatio_world} generate explorable scenes in real time from an image or a reference video.
For camera control, RealCam~\cite{realcam} and GeoStream~\cite{geostream} distill causal camera-controlled generators, conditioned on camera extrinsics and on a point cloud reprojected from the latest generated frame, respectively.
{\ourmethodar} instead conditions on the anchor rendering, which is both scale-aware and view-complete.

\vspace{1mm}
\noindent\textbf{3D reconstruction and view synthesis.}
Neural rendering methods such as NeRF~\cite{nerf} and Gaussian Splatting~\cite{gaussian_splatting} optimize a scene from many views, while feed-forward models predict radiance fields or Gaussians from one or a few images~\cite{lrm,pixelsplat,gslrm}.
Generative methods complete unobserved regions with learned priors, either by synthesizing camera-conditioned views~\cite{zero123,seva,cat3d,cat4d} or by reconstructing objects from generated multi-view images or structured 3D latents~\cite{instantmesh,lgm,trellis}.
SAM 3D Objects~\cite{sam3d} predicts the geometry, appearance and placement of an object from a single image.
{\ourmethod} reconstructs only the anchor object with SAM 3D Objects and leaves the rest of the scene, including its dynamics, to the video model.

\begin{figure*}[t]
  \centering
  \begin{subfigure}[t]{0.678\linewidth}
    \centering
    \includegraphics[width=\linewidth]{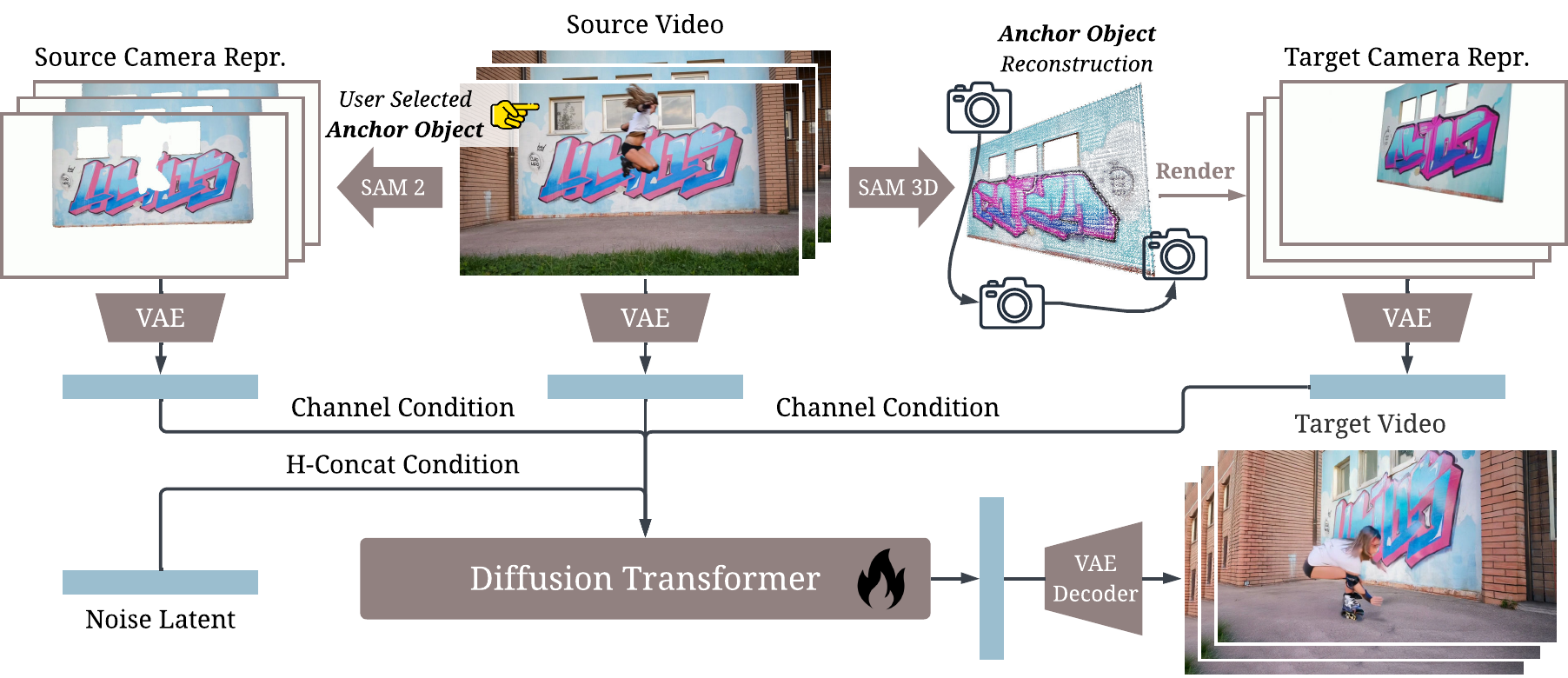}
    \subcaption{Anchor-conditioned video-to-video generation.}
    \label{fig:overview}
  \end{subfigure}\hfill
  \begin{subfigure}[t]{0.302\linewidth}
    \centering
    \includegraphics[width=\linewidth]{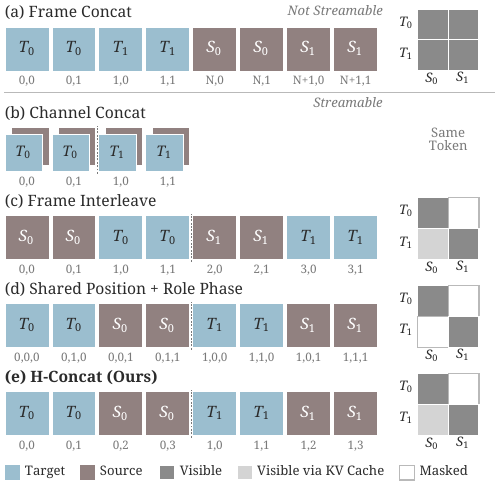}
    \subcaption{Source and target token layouts.}
    \label{fig:concat_layouts}
  \end{subfigure}
  \vspace{-2mm}
  \caption{\textbf{Overview of {\ourmethod}.}
  (a) The user selects an anchor object in the source video.
  SAM~2~\cite{sam2} tracks it to give the source camera representation (the anchor cut-out), and SAM~3D~\cite{sam3d} reconstructs it; the reconstruction is aligned to the source cameras and rendered along the target trajectory to give the target camera representation.
  The noisy target latent and the source latent are concatenated along the height (H-Concat), each half receives its camera representation through channel concatenation, and the diffusion transformer denoises the target half, which the VAE decodes into the target video.
  (b) The numbers under the tokens are their RoPE positions $(t, h)$, with the role ID as a third entry in the fourth layout. The grids mark which source frames each target frame attends to.
  Frame concatenation~\cite{recammaster, depthdirector,facecam} places the whole source after the target, so the first target frame needs the complete source video and therefore the generation cannot stream.
  Channel concatenation~\cite{gcd} merges the two frames into one token, which loses information and lowers fidelity to the source video~\cite{recammaster}.
  Frame interleaving~\cite{realcam} doubles the length of the temporal axis.
  A shared position with a role phase~\cite{joyai_video_edit} needs an additional positional encoding to tell the two streams apart.
  Our H-Concat keeps the source and target as separate tokens at the same temporal index and offsets the source in height, so the base model is used without new parameters or a custom attention mask.}
  \vspace{-4mm}
  \label{fig:method}
\end{figure*}

\vspace{-2mm}
\section{Method}
\vspace{-2mm}
\label{sec:method}
Given a source video and a target camera trajectory, {\ourmethod} generates a video that displays the same scene from the desired trajectory.
Unlike existing camera-controlled video generation methods~\cite{recammaster,redirector,trajectorycrafter,gen3c,cameractrl}, %
{\ourmethod} focuses on object-centric scenarios, which are challenging for their camera representations. 
We use a scale-aware and view-complete camera conditioning method: any object in the scene can serve as an anchor, and the rendering of its 3D reconstruction along the target trajectory provides an accurate camera signal to the video model.

\vspace{-1mm}
\subsection{Preliminaries: Scale Ambiguity}
\vspace{-1mm}
\label{sec:prelim}
Consider a video with frames $i=1,\dots,N$.
The camera of frame $i$ is described by its intrinsic matrix $K\in\mathbb{R}^{3\times3}$ and its extrinsics, where the rotation $R_i\in SO(3)$ and the translation $\mathbf{t}_i\in\mathbb{R}^3$ map world coordinates to camera coordinates.
A 3D point $\mathbf{x}\in\mathbb{R}^3$ in world coordinates projects to the pixel location $\mathbf{u}\in\mathbb{R}^2$ given by
\begin{equation}
  \mathbf{u} = \pi\big(K(R_i \mathbf{x} + \mathbf{t}_i)\big),
  \quad
  \pi\big((a, b, c)^\top\big)=(a/c,\; b/c)^\top,
  \label{eq:proj}
\end{equation}
where $\pi$ denotes the perspective division.
Since $\pi(\alpha\mathbf{y})=\pi(\mathbf{y})$ for $\alpha>0$, scaling the scene and the translation by the same factor leaves the image unchanged:
\begin{equation}
  \pi\big(K(R_i\,\alpha\mathbf{x} + \alpha\mathbf{t}_i)\big)
  = \pi\big(\alpha K(R_i \mathbf{x} + \mathbf{t}_i)\big)
  = \mathbf{u}.
  \label{eq:scale}
\end{equation}
A monocular video therefore determines the scene and the camera translation only up to a global scale.
A model conditioned directly on $[R_i\,|\,\mathbf{t}_i]$ (or on rays derived from it) must guess this undetermined scale, so the same parameter change can produce different visual transformations, leading to inaccurate camera control (\cref{fig:cam_rep}a).

FaceCam~\cite{facecam} avoids this ambiguity for portraits by conditioning on the projected facial landmarks of a generic head, whose size in the image fixes the scale.
This requires a facial landmark detector and a head prior, so it applies only while a face is visible.

\vspace{-1mm}
\subsection{Anything Can Represent the Camera}
\vspace{-1mm}
\label{sec:method_anycam}
\begin{figure}[t]
  \centering
  \includegraphics[width=0.98\linewidth]{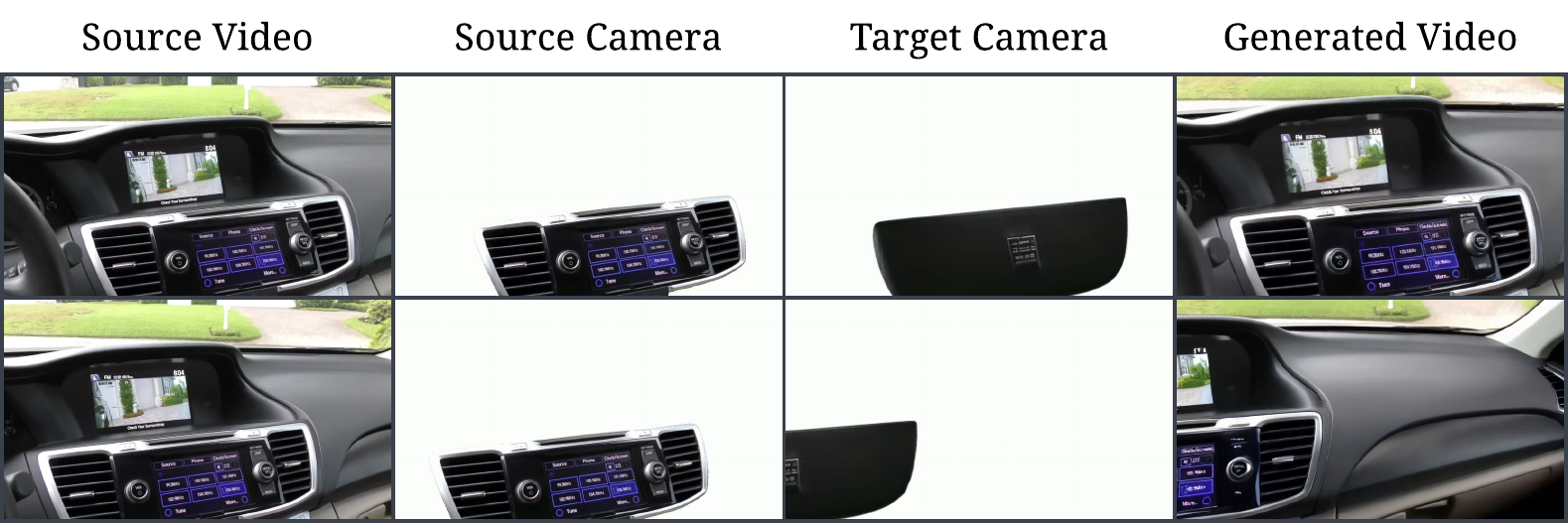}
  \vspace{-2mm}
  \caption{\textbf{Anchor rendering as a camera signal only.}
  The anchor (center console) is poorly reconstructed with no texture (column 3), compared with its cut-out in the source video (column 2).
  {\ourmethod} still recovers the appearance of the console and the cabin from the source video and controls the camera movement correctly, showing that the rendering only provides a camera signal and the model is robust to poor reconstructions.}
  \label{fig:sam3d_condition}
  \vspace{-6mm}
\end{figure}

We generalize the scale-aware conditioning of FaceCam~\cite{facecam} to general videos:
the rendering of any object (the \textit{anchor}) in the scene can serve as the camera representation (Fig.~\ref{fig:overview}).
Let $V^s=\{I^s_i\}_{i=1}^{N}$ be the source video and $\{(R_i,\mathbf{t}_i)\}_{i=1}^{N}$ the target trajectory.
We reconstruct an object visible in $V^s$ as a set of 3D Gaussians $\mathcal{G}$, place it in the coordinate system of the source cameras, and render it along the target trajectory:
\begin{equation}
  C^t_i = \mathcal{R}\big(\mathcal{G};\, K, R_i, \mathbf{t}_i\big), \qquad i=1,\dots,N,
  \label{eq:anchor_render}
\end{equation}
where $\mathcal{R}$ rasterizes the Gaussians onto a white background; $C^t$ is the target camera representation.
To indicate which object in the source video the rendering refers to, we also provide the source camera representation $C^s=\{C^s_i\}$, the anchor cut-out: the segmentation of the anchor in each source frame on a white background.
The task is then to generate the target video $V^t$ from $(V^s, C^s, C^t)$.

Three properties make this representation effective.
(1) \textit{Scale-aware.} By \cref{eq:scale}, scaling the anchor and the translation together leaves $C^t$ unchanged, and since $\mathcal{G}$ is known, the pose can be recovered from $C^t$, \eg by PnP; the anchor thus gives a scale-consistent, object-relative reference without losing camera information. The model then infers the rest of the scene relative to the anchor from the source video.
(2) \textit{View-complete.} The reconstructed object remains defined on its unobserved side, where warped point clouds or depth are empty (Fig.~\ref{fig:cam_rep}), although that side is estimated from learned priors.
(3) \textit{Object-relative.} Object-centric trajectories, such as a $90^\circ$ orbit around the anchor, can be specified directly without the metric scale of the scene.
In addition, as shown in Fig.~\ref{fig:sam3d_condition}, the model takes the appearance of the anchor, like the rest of the scene, from the source video, so an imperfect reconstruction, \eg wrong texture on the unobserved side, does not carry over to the output.

\begin{figure}[t]
  \centering
  \vspace{-2mm}
  \includegraphics[width=0.98\linewidth]{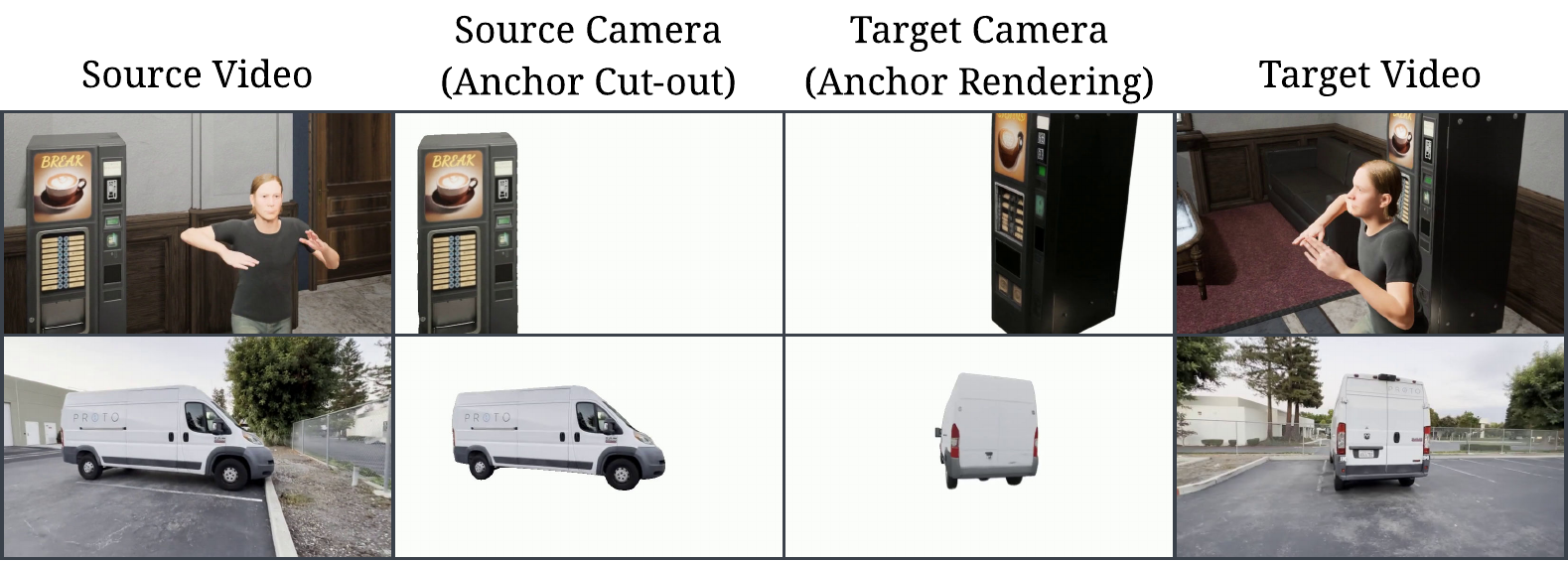}
  \vspace{-2mm}
  \caption{\textbf{Training samples.} Top: MultiCamVideo~\cite{recammaster}, a dynamic scene captured by two synchronized cameras. Bottom: DL3DV~\cite{dl3dv}, two clips of a static capture.}
  \vspace{-4mm}
  \label{fig:training_data}
\end{figure}

\vspace{-1mm}
\subsection{Training Data Construction}
\vspace{-1mm}
\label{sec:method_data}
Training requires quadruplets $(V^s, C^s, C^t, V^t)$ in which the rendering $C^t$ is pixel-aligned with the anchor in the real target video $V^t$ (Fig.~\ref{fig:training_data}), which we build from two datasets.
MultiCamVideo~\cite{recammaster} provides synthetic dynamic scenes captured by synchronized cameras, which teaches the model to keep the same dynamic content under a different trajectory, but its trajectories are simple.
DL3DV~\cite{dl3dv} provides real captures of static scenes with diverse trajectories; since the scene is static, two clips of the same capture form a synchronized source and target pair.
In total, we use 7{,}810 MultiCamVideo and 5{,}217 DL3DV samples.

\vspace{1mm}
\noindent\textbf{Anchor reconstruction.}
To build the training data, a vision-language model~\cite{qwen3vl} proposes an anchor that is visible throughout the video, rigid and stationary, large enough for reconstruction in at least one frame, and has a clear orientation; the last criterion matters because a rotationally symmetric object looks nearly the same from all sides and cannot convey rotation.
SAM~3~\cite{sam3} grounds the object and SAM~2~\cite{sam2} propagates its mask through the video, which gives $C^s$.
We reconstruct the anchor with SAM~3D Objects~\cite{sam3d} from the frame $a$ in which it is most visible, giving a Gaussian splat and its pose in the camera of that frame.

\vspace{1mm}
\noindent\textbf{Anchor alignment.}
The pose predicted by SAM~3D Objects has an unknown scale and is not accurate enough over the whole video, while the dataset cameras $\{W_i\}$ have their own world frame and scale.
We keep $\mathcal{G}$ in the camera of frame $a$, map every camera into this frame with the relative pose $W_i W_a^{-1}$, and fit the placement of $\mathcal{G}$ by render-and-compare to the SAM masks of frames that view the anchor from different sides.
We first search a single scale for the translation of all relative poses, which the parallax of the moving camera makes identifiable, and then refine a 7-DoF similarity transform of $\mathcal{G}$ with Adam. Details are provided in \cref{sec:supp_align}.
Rendering the aligned $\mathcal{G}$ with the target cameras gives $C^t$, which is pixel-aligned with the anchor in the target clip.

\vspace{-1mm}
\subsection{Model, Training and Inference}
\label{sec:method_model}
\vspace{-1mm}

\noindent\textbf{Time-synchronized concatenation (H-Concat).}
We build {\ourmethod} on the latent diffusion transformer model Wan2.2-5B~\cite{wan}.
Let $\mathbf{z}^s$, $\mathbf{z}^t$, $\mathbf{c}^s$ and $\mathbf{c}^t$ be the latents of $V^s$, $V^t$, $C^s$ and $C^t$, each of height $H$.
We concatenate the noisy target latent $\mathbf{z}^t_\sigma$ and the clean source latent $\mathbf{z}^s$ along the height, with the target on top, and give each half its camera representation, $\mathbf{c}^t$ or $\mathbf{c}^s$, through the channel-concatenated condition input of the model (Fig.~\ref{fig:overview}); we denote the result by $\mathbf{x}$.
With the 3D RoPE of the DiT, a target token at latent frame $j$ and row $h$ has position $(j, h)$ and the source token below it has $(j, h+H)$, so source and target frames of the same time step share one temporal index.
Fig.~\ref{fig:concat_layouts} compares H-Concat with other ways of combining the source and target.

\vspace{1mm}
\noindent\textbf{Training.}
Only the target half is noised, $\mathbf{z}^t_\sigma=(1-\sigma)\mathbf{z}^t+\sigma\boldsymbol{\epsilon}$ with $\boldsymbol{\epsilon}\sim\mathcal{N}(\mathbf{0},\mathbf{I})$, and supervised with the flow-matching objective of the base model:
\begin{equation}
  \mathcal{L}_{\text{FM}} = \mathbb{E}_{\sigma,\boldsymbol{\epsilon}}
  \Big\| \big[\, v_\theta(\mathbf{x}, \sigma, \tau) \,\big]_{\text{tgt}} - \big(\boldsymbol{\epsilon} - \mathbf{z}^t\big) \Big\|_2^2,
  \label{eq:fm}
\end{equation}
where $v_\theta$ is the velocity predicted by the DiT, $[\cdot]_{\text{tgt}}$ takes its target half and $\tau$ is the text prompt.
We fine-tune the self-attention layers of all DiT blocks and the input patch embedding, and keep the other weights frozen.

\vspace{1mm}
\noindent\textbf{Inference.}
At inference, ground truth cameras are not available.
We estimate cameras of the source video with ViPE~\cite{vipe} and reconstruct and align the anchor as in \cref{sec:method_data}, using the masks of the source video.
The user defines the target trajectory relative to the anchor, \eg an orbit around its center at the current camera-to-anchor distance, and rendering $\mathcal{G}$ along it gives $C^t$.
A single image input is treated as a static video by repeating it.

\vspace{-1mm}
\subsection{Streaming Generation}
\vspace{-1mm}

\label{sec:method_ar}
To adjust the anchor pose and control the camera on the fly, we train {\ourmethodar}, an autoregressive few-step camera-controlled video generator.
In the streaming setting, the anchor is placed once from the first frame with the pose predicted by SAM~3D~\cite{sam3d}, and the source anchor cut-out is tracked frame by frame with SAM~2~\cite{sam2}.

\vspace{1mm}
\noindent\textbf{Chunk-wise causal attention.}
Since H-Concat places source and target frames of the same time step at one temporal index, splitting the latent video along time gives each chunk $c$ both its source rows $S_c$, which hold $\mathbf{z}^s$ and $\mathbf{c}^s$, and its target rows $T_c$, which hold the target latent and $\mathbf{c}^t$.
For each chunk, the clean $S_c$ is processed first and appended to a key-value cache; $T_c$ is then denoised while attending to the cache, to $S_c$ and to itself; finally, the generated $T_c$ is processed at noise level $0$ and appended to the cache.
Thus, no chunk sees the source or the anchor rendering of a later chunk (Fig.~\ref{fig:concat_layouts}), and the positional encoding is the same as in the bidirectional model.
Since the VAE is causal, the source and the anchor rendering can be encoded chunk by chunk as they arrive.

\vspace{1mm}
\noindent\textbf{Teacher forcing.}
We initialize {\ourmethodar} from the bidirectional model and train it with \cref{eq:fm}, sampling the noise level of each chunk independently, while the cache holds the ground-truth target latents.

\vspace{1mm}
\noindent\textbf{Distribution matching distillation.}
Teacher forcing leaves a gap between training, where the history is ground truth, and inference, where it is generated.
Following Self Forcing~\cite{self_forcing}, we close this gap and reduce the number of sampling steps with distribution matching distillation (DMD)~\cite{dmd}, training {\ourmethodar} on its own autoregressive rollouts with a few denoising steps per chunk and the bidirectional model as the teacher.

\vspace{1mm}
\noindent\textbf{Long videos.}
Beyond the training length, each chunk attends to the first chunk, which acts as an attention sink, and to a sliding window of recent chunks, whose temporal positions are renumbered to match the layout seen in training.
To reduce the color drift that accumulates in the generated history, we append each finished chunk to the cache at a small noise level $\sigma_h$ instead of $0$, so that later chunks rely less on its exact colors.

\begin{figure*}[tb]
  \centering
  \includegraphics[width=0.98\linewidth]{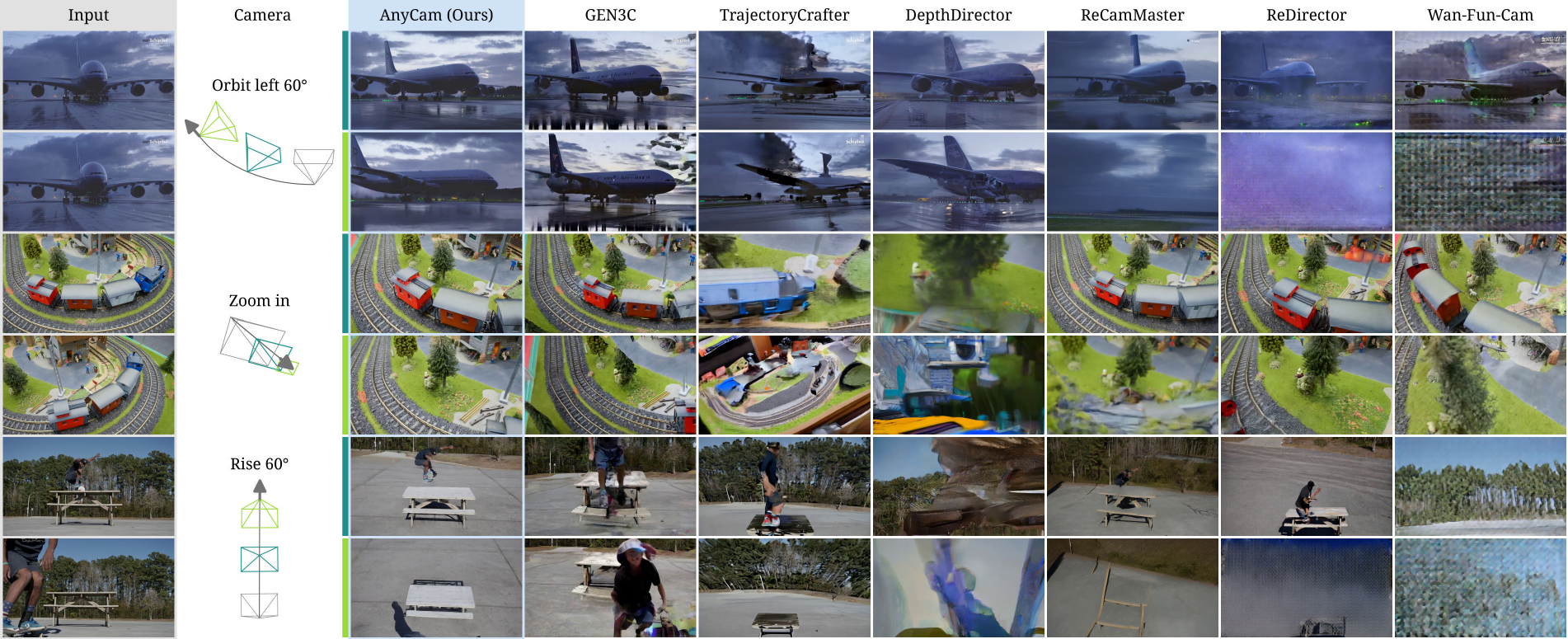}
  \vspace{-2mm}
  \caption{\textbf{Visual comparison on DAVIS.}
  {\ourmethod} follows every trajectory and keeps the scene intact, since the anchor rendering fixes the scale of the motion relative to the anchor and covers it from every side.
  Methods that warp estimated geometry show holes and stretched content at new viewpoints: TrajectoryCrafter~\cite{trajectorycrafter} tears the airplane apart during rotation (rows 1 and 2), GEN3C~\cite{gen3c} leaves stripes (row 2), and DepthDirector~\cite{depthdirector} breaks down under zoom-in and rise (rows 3 to 6).
  Methods conditioned on camera parameters guess the scale of the motion: ReCamMaster~\cite{recammaster} loses the airplane after the orbit (row 2), Wan-Fun-Cam~\cite{wan} zooms far past the train into the tree (row 4), ReDirector~\cite{redirector} and Wan-Fun-Cam collapse into noise under large orbit and rise (rows 2 and 6).}
  \vspace{-4mm}
  \label{fig:davis_qual}
\end{figure*}
 
\begin{table}[t]
  \centering
  \caption{\textbf{Video-to-video comparison on CO3D.} IoU compares the anchor tracked in the generated video with the CO3D object mask. {\ourmethod} and {\ourmethodar} generate videos more similar to the ground truth and with better visual quality and object consistency. FPS is measured on one RTX PRO 6000 Blackwell GPU.}
  \vspace{-2mm}
  \label{tab:co3d}
  \scriptsize
  \setlength{\tabcolsep}{1.2pt}
  \begin{tabular}{@{}l@{\hskip 1.5mm}c@{\hskip 2.1mm}c@{\hskip 2.1mm}c@{\hskip 1.5mm}c@{\hskip 1.5mm}c@{\hskip 2.1mm}c@{\hskip 1.5mm}c@{}}
    \toprule
    & \multicolumn{3}{c}{Ground Truth} & \multicolumn{1}{c}{Quality} & \multicolumn{2}{c}{Consistency} & \multicolumn{1}{c}{Speed} \\
    \cmidrule(r){2-4} \cmidrule(r){5-5} \cmidrule(r){6-7} \cmidrule(r){8-8}
    Method & PSNR$\uparrow$ & SSIM$\uparrow$ & LPIPS$\downarrow$ & FID$\downarrow$ & IoU$\uparrow$ & CLIP-V$\uparrow$ & FPS$\uparrow$ \\
    \midrule
    3DGS & \underline{13.76} & \textbf{0.546} & \underline{0.618} & 236.6 & \underline{0.465} & 64.76 & - \\
    GEN3C & 12.89 & 0.518 & 0.632 & 96.9 & 0.346 & 78.44 & 0.40 \\
    TrajectoryCrafter & 11.98 & 0.470 & 0.690 & 103.7 & 0.167 & 76.60 & 0.20 \\
    DepthDirector & 12.17 & \underline{0.530} & 0.676 & 155.0 & 0.174 & 67.02 & \underline{0.82} \\
    ReCamMaster & 12.55 & 0.491 & 0.631 & 75.3 & 0.273 & 79.83 & 0.17 \\
    ReDirector & 12.62 & 0.507 & 0.637 & 95.7 & 0.166 & 74.73 & 0.15 \\
    Wan-Fun-Cam & 11.74 & 0.476 & 0.666 & 82.8 & 0.282 & 83.10 & \textbf{0.96} \\
    Wan-Fun-Cam$^\dagger$ & 11.92 & 0.463 & 0.627 & \underline{74.8} & 0.343 & \underline{83.95} & 0.52 \\
    {\ourmethod} & \textbf{14.08} & \underline{0.530} & \textbf{0.500} & \textbf{56.0} & \textbf{0.814} & \textbf{86.59} & 0.57 \\
    \midrule
    HY-WorldPlay & 12.25 & 0.443 & 0.642 & 73.2 & 0.325 & 83.38 & 3.24 \\
    InSpatio-World & 12.88 & 0.481 & 0.607 & 71.0 & 0.338 & 82.10 & 3.45 \\
    Wan-Fun-Cam-AR$^\dagger$ & \underline{12.96} & \underline{0.495} & \underline{0.601} & \underline{68.9} & \underline{0.386} & \underline{85.96} & \textbf{5.42} \\
    {\ourmethodar} & \textbf{14.56} & \textbf{0.520} & \textbf{0.492} & \textbf{58.8} & \textbf{0.814} & \textbf{86.14} & \underline{3.63} \\
    \bottomrule
  \end{tabular}
  \vspace{-4mm}
\end{table}

\vspace{-1mm}
\section{Experiments}
\label{sec:experiments}
\vspace{-1mm}
\subsection{Experimental Setup}
\label{sec:exp_setup}
\vspace{-1mm}

\noindent\textbf{Implementation details.}
We build {\ourmethod} on Wan2.2-Fun-5B-InP~\cite{wan} and train it on the data described in~\cref{sec:method_data} for 10K steps with AdamW, learning rate $5\times10^{-5}$, batch size 32. Inference uses 50 sampling steps and text guidance of scale 5 and generates videos with 81 frames at $832\times480$.
{\ourmethodar} is built on the bidirectional model and trained with LoRA~\cite{lora} of rank 128, using chunks of 3 latent frames.
Anchors are selected with Qwen3-VL~\cite{qwen3vl}.

\vspace{1mm}
\noindent\textbf{Datasets and metrics.}
(1) \textit{DAVIS}~\cite{davis_2016,davis_2017}.
We select 50 dynamic videos and apply 9 object-centric trajectories around the anchor, adapted from the basic trajectories of ReCamMaster~\cite{recammaster}. %
Since no ground truth is available, we estimate camera poses of the generated videos with ViPE~\cite{vipe} and report rotation error (Rot.), translation error (Trans.) and summed error of the camera-to-world matrices (CamMC).
We also report visual quality with FID~\cite{fid} and VBench~\cite{vbench}.
(2) \textit{CO3D}~\cite{co3d}.
We take 34 object-centric sequences and build three target arcs (easy, medium, and hard) for each, %
yielding 102 cases with ground-truth frames.
(3) \textit{Mip-NeRF 360}~\cite{mipnerf360}.
In each of 9 scenes, 4 target trajectories start from the single input photo and pass through other photos of the scene, which serve as the ground truth, giving 36 cases in total.
(4) \textit{RealEstate10K}~\cite{re10k}.
We select 100 clips to evaluate long video generation. %
On CO3D, Mip-NeRF 360 and RealEstate10K, where ground truth is available, we report PSNR, SSIM and LPIPS~\cite{lpips}.

\vspace{1mm}
\noindent\textbf{Baselines.}
We compare with camera-controlled video generation methods: GEN3C~\cite{gen3c}, TrajectoryCrafter~\cite{trajectorycrafter}, DepthDirector~\cite{depthdirector}, ReCamMaster~\cite{recammaster}, ReDirector~\cite{redirector}, and Wan2.2-Fun-Control-Camera (Wan-Fun-Cam)~\cite{wan}.
We also compare with HY-WorldPlay~\cite{worldplay} and InSpatio-World~\cite{inspatio_world} for autoregressive generation.
To evaluate NVS, we compare with 3DGS~\cite{gaussian_splatting} optimized on 81 frames, and with OpenLRM~\cite{lrm}, LGM~\cite{lgm}, TRELLIS~\cite{trellis}, SAM~3D~\cite{sam3d} and SEVA~\cite{seva} in the single-image setting.

\begin{figure*}[t]
  \centering
  \includegraphics[width=0.98\linewidth]{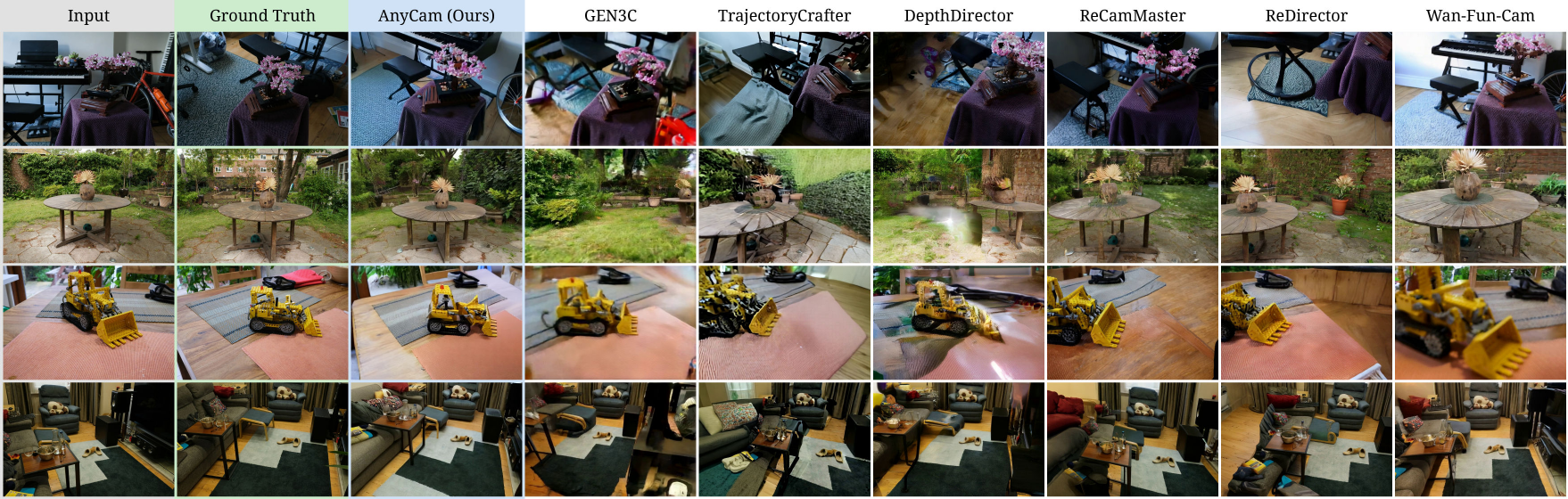}
  \vspace{-2mm}
  \caption{\textbf{Single-image to video comparison on Mip-NeRF 360.}
  Since the anchor rendering is object-relative and view-complete, {\ourmethod} always reaches the target viewpoints.
  Wan-Fun-Cam~\cite{wan} and ReCamMaster~\cite{recammaster} are conditioned on camera parameters, whose scale a single photo cannot determine, and barely move away from the input view (rows 1 to 3).
  GEN3C~\cite{gen3c}, TrajectoryCrafter~\cite{trajectorycrafter} and DepthDirector~\cite{depthdirector} warp a depth map that covers only the visible surface, which leaves distorted or missing content (rows 1 to 3).}
  \label{fig:mip_qual}
  \vspace{-4mm}
\end{figure*}

\begin{table}[t]
  \centering
  \caption{\textbf{Video-to-video comparison on DAVIS.} {\ourmethod} has the lowest translation error and the best visual quality and consistency; although GEN3C~\cite{gen3c} and ReDirector~\cite{redirector} achieve better rotation accuracy, their visual quality suffers.}
  \vspace{-2mm}
  \label{tab:davis_bidi}
  \scriptsize   %
  \setlength{\tabcolsep}{1pt}
  \begin{tabular}{@{}l@{\hskip 1.5mm}c@{\hskip 1.5mm}c@{\hskip 1.5mm}c@{\hskip 1.5mm}c@{\hskip 1.5mm}c@{\hskip 1.5mm}c@{\hskip 1.5mm}c@{}}
    \toprule
    & \multicolumn{3}{c}{Camera} & \multicolumn{2}{c}{Quality} & \multicolumn{2}{c}{Consistency} \\
    \cmidrule(r){2-4} \cmidrule(r){5-6} \cmidrule(r){7-8}
    Method & Rot.$\downarrow$ & Trans.$\downarrow$ & CamMC$\downarrow$ & FID$\downarrow$ & VBench$\uparrow$ & CLIP-V$\uparrow$ & CLIP-F$\uparrow$ \\
    \midrule
    GEN3C & \textbf{3.78} & \underline{0.046} & \textbf{0.138} & 60.72 & 79.56 & 82.76 & 96.53 \\
    TrajectoryCrafter & 13.48 & 0.086 & 0.334 & 45.85 & 77.40 & 83.89 & 94.97 \\
    DepthDirector & 10.48 & 0.119 & 0.337 & 82.78 & 75.25 & 75.66 & 94.84 \\
    ReCamMaster & 4.94 & 0.054 & 0.182 & \underline{41.60} & \underline{81.94} & \underline{87.58} & 96.84 \\
    ReDirector & \underline{4.10} & 0.055 & \underline{0.156} & 47.28 & 81.36 & 85.70 & 96.83 \\
    Wan-Fun-Cam & 6.90 & 0.066 & 0.237 & 58.61 & 80.72 & 83.92 & \underline{97.32} \\
    Wan-Fun-Cam$^\dagger$ & 4.83 & 0.057 & 0.179 & 50.74 & 80.77 & 85.69 & 96.80 \\
    {\ourmethod} & 4.60 & \textbf{0.028} & 0.159 & \textbf{41.26} & \textbf{82.89} & \textbf{88.53} & \textbf{97.34} \\
    \bottomrule
  \end{tabular}
  \vspace{-4mm}
\end{table}

\vspace{-1mm}
\subsection{Video-to-Video Camera Control}
\label{sec:exp_v2v}
\vspace{-1mm}

\noindent\textbf{CO3D.}
As shown in \cref{tab:co3d}, {\ourmethod} has the best PSNR, LPIPS and FID, and its anchor IoU of 0.814 is more than twice that of every video generation baseline. %
The anchor rendering is view-complete (\cref{sec:method_anycam}), so it still conveys the camera when the target reaches the unobserved side.
3DGS~\cite{gaussian_splatting} has the highest SSIM because it reproduces the observed surfaces, but it breaks down on the unobserved side (FID 236.6).
\Cref{fig:co3d_qual} in the appendix shows qualitative results.
In the autoregressive group, {\ourmethodar} outperforms the baselines on every quality metric.
It streams at 3.63 FPS, faster than the baselines~\cite{worldplay,inspatio_world}.

\vspace{1mm}
\noindent\textbf{DAVIS.}
As shown in \cref{tab:davis_bidi}, {\ourmethod} has the lowest translation error, 0.028 against 0.046 for the next best method.
Since a monocular video does not determine the scene scale, camera parameters leave the length of a translation ambiguous, while the size of the anchor in the rendering shows it directly.
GEN3C~\cite{gen3c} and ReDirector~\cite{redirector} have lower rotation errors, but their frames degrade under large motions (\cref{fig:davis_qual}), which leads to a higher FID.
We also evaluate consistency with CLIP-V, the CLIP~\cite{clip} similarity between the source and the generated frame at the same time, and CLIP-F, the similarity between consecutive generated frames.
{\ourmethod} is best on both. Its videos stay faithful to the content of source video and are temporally consistent.

\begin{table}[t]
  \centering
  \caption{\textbf{Single-image comparison on CO3D.} oPSNR, oSSIM and oLPIPS are computed inside the object mask. {\ourmethod} achieves better reconstruction quality than 3D reconstruction methods (top) and video generation methods (bottom).}
  \vspace{-2mm}
  \label{tab:co3d_i2v}
  \scriptsize
  \setlength{\tabcolsep}{4pt}
  \begin{tabular}{@{}l@{\hskip 5.5mm}c@{\hskip 5.5mm}c@{\hskip 5.5mm}c@{\hskip 5.5mm}c@{}}
    \toprule
    Method & oPSNR$\uparrow$ & oSSIM$\uparrow$ & oLPIPS$\downarrow$ & Anchor IoU$\uparrow$ \\
    \midrule
    OpenLRM & 9.59 & \textbf{0.416} & 0.513 & 0.631 \\
    LGM & 9.75 & 0.399 & 0.522 & 0.634 \\
    TRELLIS & 10.05 & 0.389 & 0.488 & 0.710 \\
    SAM~3D & 9.98 & 0.388 & 0.494 & 0.724 \\
    \midrule
    GEN3C & 10.68 & 0.358 & 0.583 & 0.303 \\
    TrajectoryCrafter & 10.43 & 0.343 & 0.605 & 0.199 \\
    DepthDirector & 10.61 & 0.365 & 0.607 & 0.184 \\
    ReCamMaster & 10.53 & 0.355 & 0.609 & 0.286 \\
    ReDirector & 10.27 & 0.351 & 0.629 & 0.151 \\
    SEVA & 12.12 & 0.356 & 0.483 & 0.590 \\
    HY-WorldPlay & 10.16 & 0.308 & 0.590 & 0.330 \\
    InSpatio-World & 11.09 & 0.345 & 0.568 & 0.351 \\
    Wan-Fun-Cam & 10.69 & 0.353 & 0.584 & 0.272 \\
    Wan-Fun-Cam$^\dagger$ & 10.59 & 0.354 & 0.559 & 0.399 \\
    Wan-Fun-Cam-AR$^\dagger$ & 11.12 & 0.356 & 0.547 & 0.413 \\
    {\ourmethodar} & \underline{13.26} & 0.385 & \underline{0.416} & \underline{0.731} \\
    {\ourmethod} & \textbf{13.52} & \underline{0.410} & \textbf{0.401} & \textbf{0.736} \\
    \bottomrule
  \end{tabular}
  \vspace{-2mm}
\end{table}

\begin{table}[t]
  \centering
  \caption{\textbf{Single-image comparison on Mip-NeRF 360.} {\ourmethod} is best on all ground-truth metrics and FID.}%
  \vspace{-2mm}
  \label{tab:mip}
  \scriptsize
  \setlength{\tabcolsep}{2pt}
  \begin{tabular}{@{}l@{\hskip 5.5mm}c@{\hskip 5.5mm}c@{\hskip 5.5mm}c@{\hskip 5.5mm}c@{\hskip 5.5mm}c@{}}
    \toprule
    & \multicolumn{3}{c}{Ground Truth} & \multicolumn{2}{c}{Quality} \\
    \cmidrule(r){2-4} \cmidrule(r){5-6}
    Method & PSNR$\uparrow$ & SSIM$\uparrow$ & LPIPS$\downarrow$ & FID$\downarrow$ & VBench$\uparrow$ \\
    \midrule
    GEN3C & 10.91 & \underline{0.203} & 0.717 & 114.3 & 77.03 \\
    TrajectoryCrafter & 10.46 & 0.174 & 0.710 & 118.7 & 73.74 \\
    DepthDirector & \underline{11.26} & 0.193 & 0.680 & 119.7 & 74.77 \\
    ReCamMaster & 11.23 & 0.192 & \underline{0.667} & 94.8 & \textbf{81.93} \\
    ReDirector & 11.06 & 0.190 & 0.675 & 115.3 & 77.20 \\
    Wan-Fun-Cam & 10.36 & 0.182 & 0.722 & 111.4 & 76.53 \\
    Wan-Fun-Cam$^\dagger$ & 9.78 & 0.176 & 0.715 & \underline{94.3} & 77.14 \\
    {\ourmethod} & \textbf{11.86} & \textbf{0.207} & \textbf{0.585} & \textbf{80.1} & \underline{79.85} \\
    \bottomrule
  \end{tabular}
  \vspace{-4mm}
\end{table}

\begin{figure*}[t]
  \centering
  \includegraphics[width=0.98\linewidth]{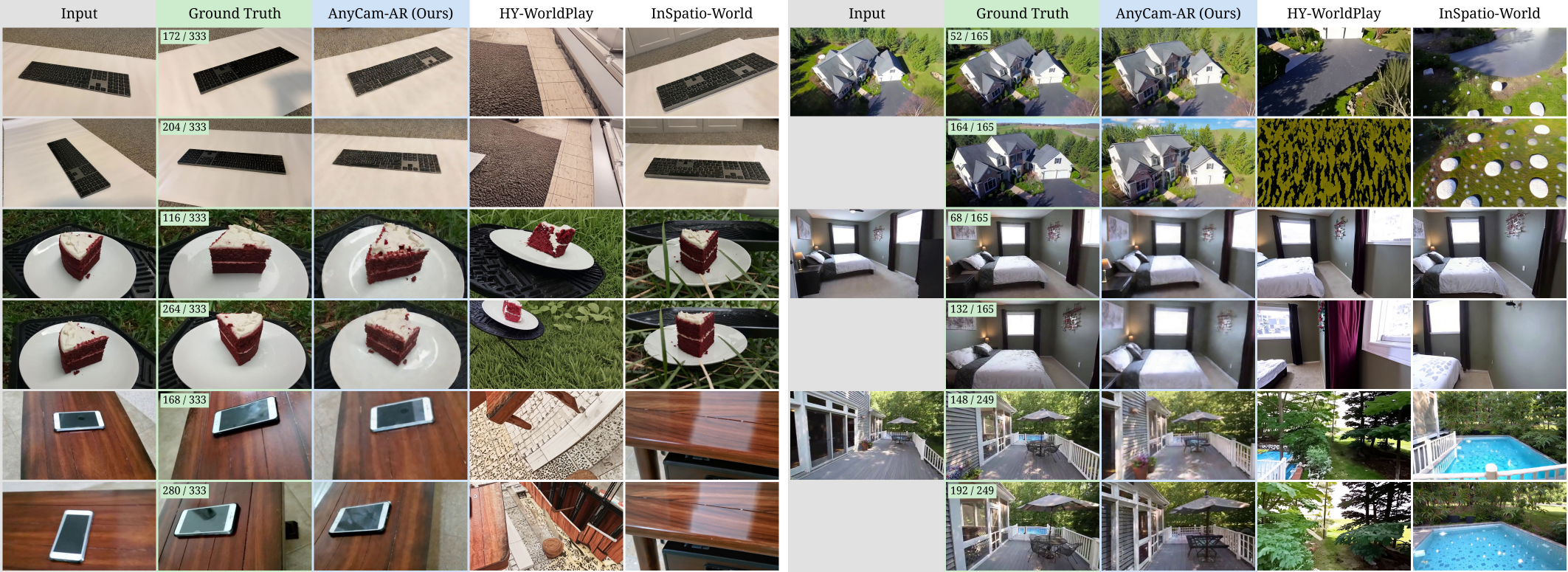}
  \vspace{-2mm}
  \caption{\textbf{Long-video comparison on CO3D (left) and RealEstate10K (right).} %
  The anchor rendering gives every chunk a reference to the object, so {\ourmethodar} stays registered to the scene over hundreds of frames.
  The baselines drift as their errors accumulate over the chunks: both lose the cellphone (left, rows 5, 6), and both turn away from the house (right, rows 1, 2) and the deck (right, rows 5, 6).}
  \label{fig:ar_qual}
  \vspace{-4mm}
\end{figure*}

\vspace{-1mm}
\subsection{Single-Image Input}
\label{sec:exp_image}
\vspace{-1mm}

\noindent\textbf{CO3D.}
Every method receives only the first source frame of each of the 102 CO3D cases.
\cref{tab:co3d_i2v} shows that the reconstruction methods recover the shape of the object but not its appearance inside the object mask.
{\ourmethod} has the best oPSNR, oLPIPS and anchor IoU, and {\ourmethodar} is second on all three.
The comparison between {\ourmethod} and SAM~3D~\cite{sam3d} shows that the reconstruction quality of the anchor object does not bound the output quality of {\ourmethod}.

\vspace{1mm}
\noindent\textbf{Mip-NeRF 360.}
\cref{tab:mip} shows that {\ourmethod} has the best PSNR, SSIM, LPIPS and FID.
ReCamMaster~\cite{recammaster} has the highest VBench, which rates the frames themselves but not whether they reach the target viewpoint.
The anchor rendering is object-relative and view-complete, so {\ourmethod} reaches the viewpoints of the ground-truth photos (\cref{fig:mip_qual}).

\begin{table}[t]
  \centering
  \caption{\textbf{Long-video comparison on RealEstate10K.} {\ourmethodar}'s rotation error is less than half that of InSpatio-World~\cite{inspatio_world}.}
  \vspace{-2mm}
  \label{tab:re10k}
  \scriptsize
  \setlength{\tabcolsep}{1.2pt}
  \begin{tabular}{@{}l@{\hskip 1.1mm}c@{\hskip 1.1mm}c@{\hskip 1.1mm}c@{\hskip 1.1mm}c@{\hskip 1.1mm}c@{\hskip 1.1mm}c@{}}
    \toprule
    & \multicolumn{3}{c}{Ground Truth} & \multicolumn{2}{c}{Camera} & \multicolumn{1}{c}{Anchor} \\
    \cmidrule(r){2-4} \cmidrule(r){5-6} \cmidrule(r){7-7}
    Method & mPSNR$\uparrow$ & mSSIM$\uparrow$ & mLPIPS$\downarrow$ & Pose err.$\downarrow$ & Rot.\ err.$\downarrow$ & IoU$\uparrow$ \\
    \midrule
    \multicolumn{7}{l}{\textit{165 frames}} \\
    \midrule
    HY-WorldPlay & 11.16 & 0.590 & 0.530 & 25.84 & 32.44 & 0.144 \\
    InSpatio-World & \underline{13.99} & \underline{0.654} & \underline{0.385} & \underline{10.35} & \underline{13.92} & \underline{0.433} \\
    Wan-Fun-Cam-AR$^\dagger$ & 12.07 & 0.608 & 0.529 & 18.98 & 21.14 & 0.183 \\
    {\ourmethodar} & \textbf{14.40} & \textbf{0.665} & \textbf{0.370} & \textbf{7.87} & \textbf{4.84} & \textbf{0.668} \\
    \midrule
    \multicolumn{7}{l}{\textit{249 frames}} \\
    \midrule
    HY-WorldPlay & 10.69 & 0.618 & 0.554 & 42.10 & 47.22 & 0.105 \\
    InSpatio-World & \underline{13.00} & \underline{0.666} & \underline{0.436} & \underline{21.78} & \underline{24.77} & \underline{0.338} \\
    Wan-Fun-Cam-AR$^\dagger$ & 11.73 & 0.630 & 0.540 & 26.54 & 28.15 & 0.145 \\
    {\ourmethodar} & \textbf{13.66} & \textbf{0.677} & \textbf{0.404} & \textbf{13.41} & \textbf{9.27} & \textbf{0.595} \\
    \midrule
    \multicolumn{7}{l}{\textit{417 frames, out and back}} \\
    \midrule
    HY-WorldPlay & 10.13 & 0.569 & 0.592 & 54.16 & 58.86 & 0.063 \\
    InSpatio-World & \underline{12.99} & \underline{0.648} & \underline{0.444} & \underline{25.31} & \underline{26.56} & \underline{0.309} \\
    Wan-Fun-Cam-AR$^\dagger$ & 11.75 & 0.608 & 0.551 & 26.54 & 27.93 & 0.120 \\
    {\ourmethodar} & \textbf{13.24} & \textbf{0.649} & \textbf{0.432} & \textbf{14.73} & \textbf{10.94} & \textbf{0.516} \\
    \bottomrule
  \end{tabular}
  \vspace{-4mm}
\end{table}

\subsection{Long-Video Autoregressive Generation}
\label{sec:exp_long}

\noindent\textbf{CO3D.}
As shown in~\cref{fig:ar_qual}, {\ourmethodar} keeps the anchor on the target orbit over hundreds of frames, while the baselines lose the object or drift to other parts of the scene.
Since H-Concat gives every target chunk the anchor rendering of the same time step, each chunk is told where the object is, and the errors do not accumulate as the video grows.

\vspace{1mm}
\noindent\textbf{RealEstate10K.} From a single input image, the target follows the real camera path for 165, 249, and 417 frames (208 frames and back).
mPSNR, mSSIM and mLPIPS are computed only in the part of each ground-truth frame that is visible in the input frame.
Each generated frame is registered to the ground-truth frame at the same time with MASt3R~\cite{mast3r}; Pose err.\ and Rot.\ err.\ are the rotation errors in degrees of the PnP pose and of a rotation-only fit, with unregistered frames counted as $90^\circ$.
\cref{tab:re10k} shows that {\ourmethodar} generates the frames closest to the ground truth, and its camera control stays accurate even after 417 generated frames.

\subsection{Ablation Study}
\label{sec:exp_ablation}
\vspace{-1mm}
To show that the favorable performance of {\ourmethod} and {\ourmethodar} comes from the anchor rendering rather than from the training data, we train Wan-Fun-Cam~\cite{wan} with its original Plücker-ray conditioning on our training data (\cref{sec:method_data}), which gives Wan-Fun-Cam$^\dagger$, and distill it into Wan-Fun-Cam-AR$^\dagger$.
The training recipes are the same as ours.
Although Wan-Fun-Cam$^\dagger$ improves over Wan-Fun-Cam on DAVIS and CO3D, its camera control accuracy and video quality are still lower than those of {\ourmethod} (\cref{tab:davis_bidi,tab:co3d}).
The same holds for Wan-Fun-Cam-AR$^\dagger$ and {\ourmethodar} in autoregressive generation, as shown in \cref{tab:re10k}.
This shows that the anchor rendering is the key to accurate camera control, while camera parameters remain scale-ambiguous even with the same training data.

\section{Conclusion}
\label{sec:conclusion}
\vspace{-1mm}
We introduce {\ourmethod}, a camera-controlled video generation method that represents the camera with the rendering of an anchor object.
The representation is scale-aware, view-complete and object-relative, enabling object-centric camera control.
We train {\ourmethod} on source and target pairs with pixel-aligned anchor renderings, built from synchronized multi-camera dynamic videos and static-scene captures.
With time-synchronized H-Concat layout, our model can be distilled into {\ourmethodar}, a few-step autoregressive generator that produces the target video chunk by chunk from a streaming input.
Experiments on DAVIS~\cite{davis_2016,davis_2017}, Mip-NeRF 360~\cite{mipnerf360}, CO3D~\cite{co3d} and RealEstate10K~\cite{re10k} show that {\ourmethod} performs favorably against existing camera-controlled video generation methods and reconstruction methods in camera accuracy and visual quality.

\clearpage

\section*{LLM Usage Statement}
We used Claude Code (Anthropic), an LLM-based coding assistant, to assist with literature searches and code debugging during research, as well as grammar checking and typo correction during manuscript preparation. Separately, Qwen3-VL~\cite{qwen3vl}, a vision-language model, was used as a component of our proposed method and was not involved in manuscript preparation.
All AI-generated or AI-assisted text and code were carefully reviewed and verified by the authors. The research ideas, methodology, experimental design, and conclusions were developed independently by the authors, who take full responsibility for the content of this work.

{
    \small
    \bibliographystyle{ieeenat_fullname}
    \bibliography{main}
}

\clearpage
\appendix
\clearpage
\begin{center}\Large\textbf{Appendix}\end{center}
\vspace{2mm}

\noindent This appendix includes additional experimental results (\cref{sec:supp_results}), a discussion of which objects can serve as anchors, anchor switching, failure cases and limitations (\cref{sec:supp_discussion}), implementation details of the method (\cref{sec:supp_impl}), and evaluation details (\cref{sec:supp_evaldetails}).

\vspace{1mm}
\noindent\textbf{Video.}
Our video\footnote{\url{https://youtu.be/vWr0S4sPfqw}}, also available on the project page, explains the anchor rendering and presents video comparisons with the baselines on DAVIS~\cite{davis_2016, davis_2017}, CO3D~\cite{co3d}, Mip-NeRF 360~\cite{mipnerf360}, visualizations of the camera trajectories of all comparisons, demonstrations of camera control through different objects in the same video, anchor switching when the anchor leaves the view, and an interactive demo of {\ourmethodar}.

\section{Additional Results}
\label{sec:supp_results}
\begin{figure}[t]
  \centering
  \includegraphics[width=\linewidth]{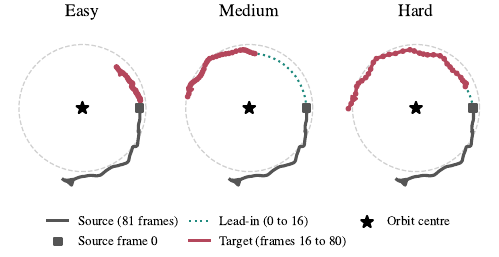}
  \vspace{-5mm}
  \caption{\textbf{CO3D target trajectories of the three difficulty levels} for one sequence (hydrant), seen from above.
  Positions are normalized by the median radius of the source cameras.
  The target camera moves from source frame 0 to the first real target camera (dotted) and then follows real CO3D cameras, with ground truth at the dots.}
  \label{fig:co3d_tiers}
\end{figure}

\begin{figure*}[t]
  \centering
  \includegraphics[width=\linewidth]{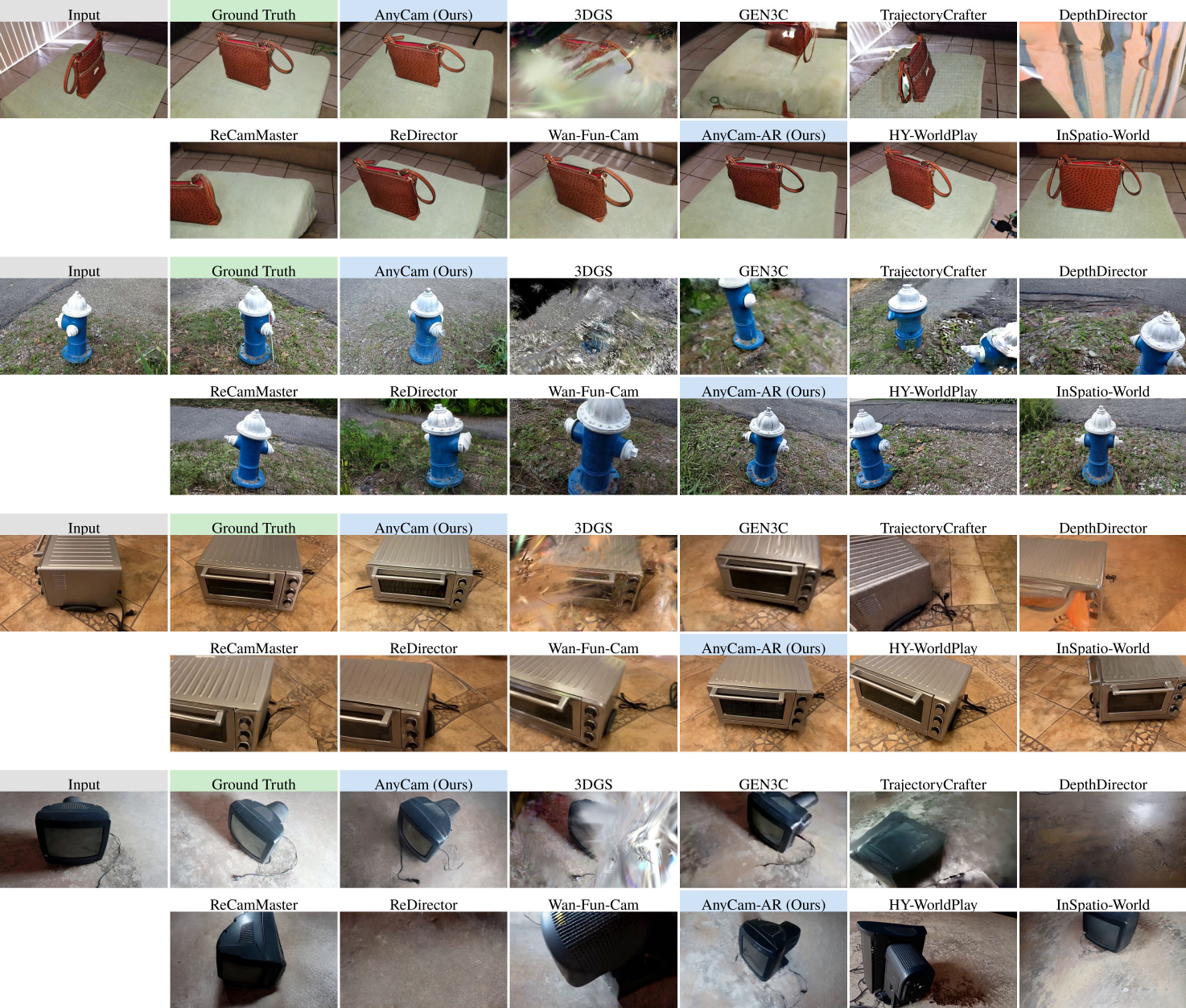}
  \caption{\textbf{Video-to-video comparison on CO3D.}
  Each case takes two rows with the source frame (Input), the ground-truth frame and the frame of every method at the same time.
  3DGS~\cite{gaussian_splatting} is optimized per scene on the 81 source frames. {\ourmethodar}, HY-WorldPlay~\cite{worldplay} and InSpatio-World~\cite{inspatio_world} are autoregressive models.
  On the sides of the objects that the source video does not show, {\ourmethod} and {\ourmethodar} match the ground truth, while 3DGS and DepthDirector~\cite{depthdirector} break down and most other baselines show the object from a wrong angle or lose it.
  For the television (bottom), ReDirector~\cite{redirector} and DepthDirector show only the floor and HY-WorldPlay shows its back.}
  \label{fig:co3d_qual}
\end{figure*}

\begin{figure*}[t]
  \centering
  \includegraphics[width=\linewidth]{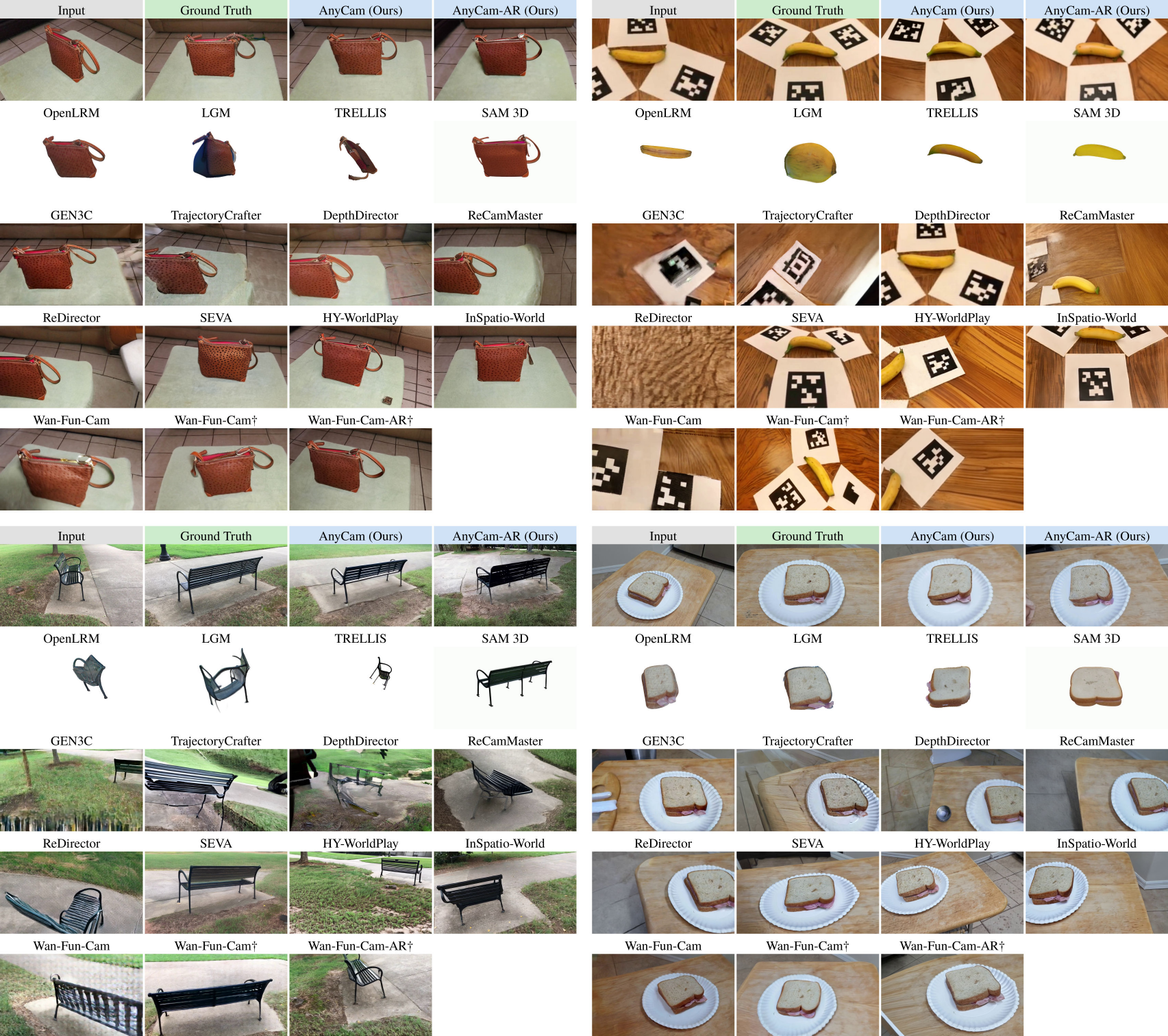}
  \caption{\textbf{Single-image comparison on CO3D.}
  Each block shows one case with the input image (Input), the ground-truth frame and the frame generated by each method of \cref{tab:co3d_i2v} from the same target camera.
  The 3D reconstruction methods render only the object and often distort its shape, \eg, LGM~\cite{lgm} for three of the four objects and TRELLIS~\cite{trellis} for the handbag and the bench.
  Most video generation baselines move to an incorrect viewpoint. GEN3C~\cite{gen3c}, TrajectoryCrafter~\cite{trajectorycrafter}, ReDirector~\cite{redirector} and Wan-Fun-Cam~\cite{wan} lose the banana (top right).
  GEN3C, TrajectoryCrafter, DepthDirector~\cite{depthdirector}, ReCamMaster~\cite{recammaster} and ReDirector shift the sandwich (bottom right) to the right.
  {\ourmethod} and {\ourmethodar} match the viewpoint and the appearance of the ground truth.}
  \label{fig:co3d_i2v_qual}
\end{figure*}

\noindent\textbf{CO3D.}
\Cref{fig:co3d_tiers} shows how the three difficulty levels of the CO3D~\cite{co3d} comparison are built for one sequence.
The source video is 81 consecutive frames of a CO3D sequence.
We measure the azimuth of every camera of the sequence about the object, using the point with the smallest sum of squared distances to all optical axes as the center and the plane fitted to the camera centers as the ground plane.
Each target trajectory starts at the camera of source frame 0 and moves along this orbit to the first target camera within 16 frames.
Frames 16 to 80 then follow real CO3D cameras outside the source window, which provide the ground truth on the even frames, and the odd frames are interpolated between them.
The levels differ in where these cameras lie relative to the source arc, \ie the azimuth range covered by the source cameras.
For easy targets, 33 consecutive cameras start next to source frame 0 and move away from the source arc, with the center of the target arc at most $45^\circ$ from the nearest source camera.
For medium targets, 33 consecutive cameras lie on the far side of the object: the center of the target arc is at least $90^\circ$ from the nearest source camera, and no target camera is within $45^\circ$ of a source camera.
For hard targets, every second or third camera is taken, so that the 33 cameras sweep $150^\circ$ to $200^\circ$ of azimuth outside the source arc, and the far end of the sweep is as close as possible to the side opposite source frame 0.
\Cref{tab:co3d_levels} splits the CO3D comparison of \cref{tab:co3d} by target difficulty, and \cref{fig:co3d_qual} shows qualitative results, including views of the unobserved side of the object.
\Cref{fig:co3d_i2v_qual} shows the single-image comparison of \cref{tab:co3d_i2v} with all baselines.
\begin{table*}[t]
  \centering
  \caption{\textbf{Video-to-video comparison on CO3D by target difficulty.}
  Easy targets have an arc center within $45^\circ$ of the nearest source camera; medium targets have an arc center at least $90^\circ$ away; and hard targets sweep $150^\circ$ to $200^\circ$ across the unobserved side of the object (34 cases each).
  IoU compares the anchor tracked in the generated video with the CO3D object mask.}
  \label{tab:co3d_levels}
  \footnotesize
  \setlength{\tabcolsep}{2pt}
  \begin{tabular*}{\textwidth}{@{\extracolsep{\fill}}lccccccccc}
    \toprule
    & \multicolumn{3}{c}{Easy} & \multicolumn{3}{c}{Medium} & \multicolumn{3}{c}{Hard} \\
    \cmidrule(r){2-4} \cmidrule(r){5-7} \cmidrule(r){8-10}
    Method & PSNR$\uparrow$ & LPIPS$\downarrow$ & IoU$\uparrow$ & PSNR$\uparrow$ & LPIPS$\downarrow$ & IoU$\uparrow$ & PSNR$\uparrow$ & LPIPS$\downarrow$ & IoU$\uparrow$ \\
    \midrule
    3DGS & \textbf{15.14} & \underline{0.550} & \underline{0.539} & \underline{12.72} & 0.668 & \underline{0.456} & \underline{13.41} & \underline{0.634} & \underline{0.399} \\
    GEN3C & 13.47 & 0.588 & 0.425 & 12.41 & 0.652 & 0.323 & 12.77 & 0.657 & 0.289 \\
    TrajectoryCrafter & 11.79 & 0.685 & 0.220 & 12.04 & 0.696 & 0.184 & 12.13 & 0.688 & 0.095 \\
    DepthDirector & 12.53 & 0.653 & 0.244 & 11.93 & 0.688 & 0.149 & 12.06 & 0.687 & 0.129 \\
    ReCamMaster & 13.18 & 0.597 & 0.251 & 12.15 & 0.654 & 0.306 & 12.32 & 0.642 & 0.264 \\
    ReDirector & 13.12 & 0.586 & 0.244 & 12.32 & 0.672 & 0.156 & 12.41 & 0.654 & 0.098 \\
    Wan-Fun-Cam & 13.07 & 0.595 & 0.361 & 10.79 & 0.713 & 0.251 & 11.37 & 0.691 & 0.232 \\
    Wan-Fun-Cam$^\dagger$ & 12.82 & 0.585 & 0.385 & 11.48 & \underline{0.649} & 0.379 & 11.46 & 0.648 & 0.265 \\
    {\ourmethod} & \underline{14.92} & \textbf{0.459} & \textbf{0.836} & \textbf{13.49} & \textbf{0.527} & \textbf{0.824} & \textbf{13.83} & \textbf{0.515} & \textbf{0.783} \\
    \midrule
    HY-WorldPlay & 13.01 & 0.596 & 0.389 & 11.76 & 0.674 & 0.308 & 11.97 & 0.656 & 0.277 \\
    InSpatio-World & \underline{13.90} & \underline{0.545} & \underline{0.420} & 12.04 & 0.653 & 0.341 & \underline{12.70} & 0.624 & 0.253 \\
    Wan-Fun-Cam-AR$^\dagger$ & 13.77 & 0.548 & 0.412 & \underline{12.41} & \underline{0.636} & \underline{0.448} & \underline{12.70} & \underline{0.618} & \underline{0.297} \\
    {\ourmethodar} & \textbf{15.47} & \textbf{0.439} & \textbf{0.834} & \textbf{13.87} & \textbf{0.532} & \textbf{0.818} & \textbf{14.34} & \textbf{0.505} & \textbf{0.791} \\
    \bottomrule
  \end{tabular*}
\end{table*}

\begin{figure*}[t]
  \centering
  \includegraphics[width=0.98\linewidth]{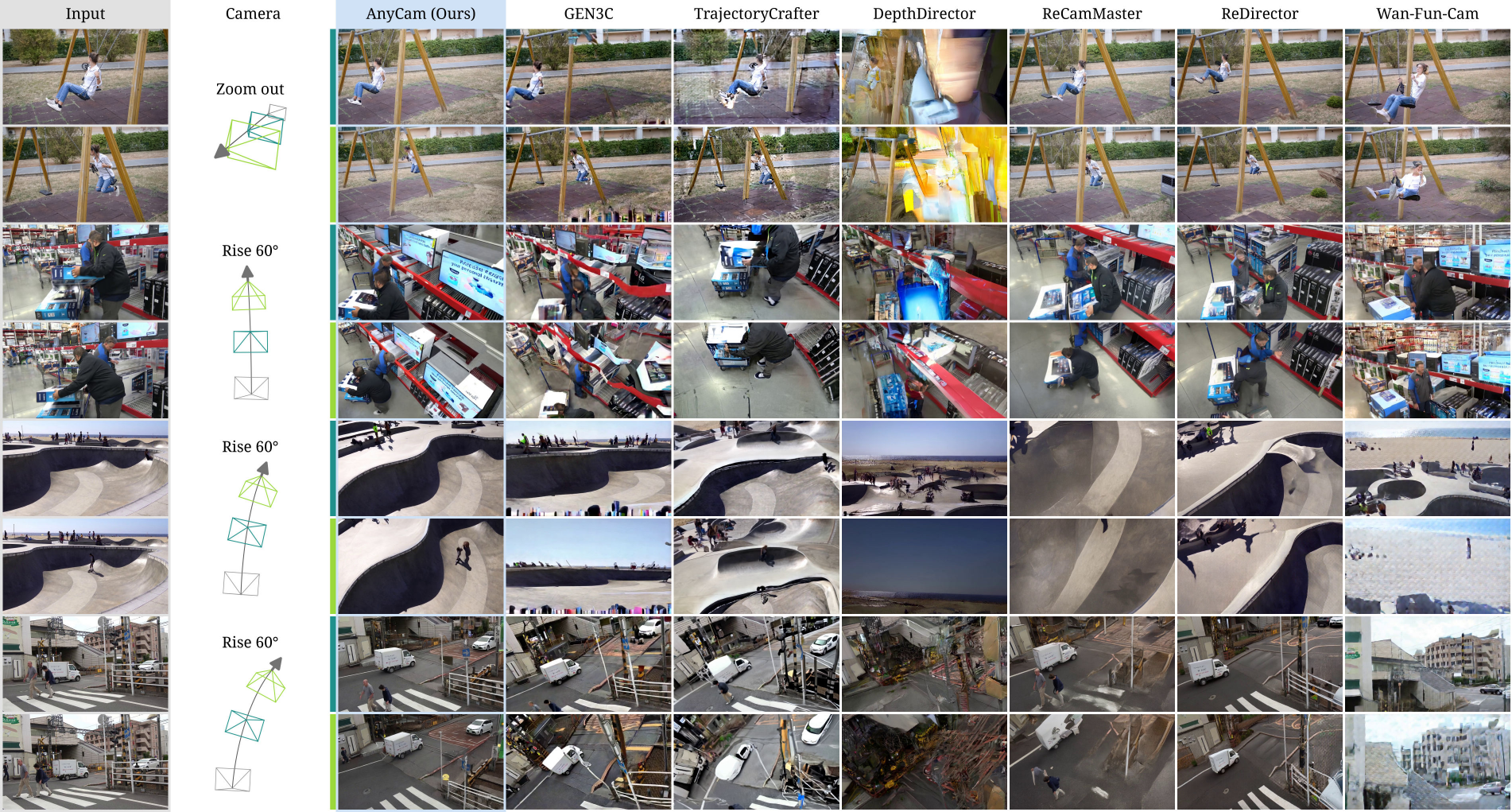}
  \vspace{-2mm}
  \caption{\textbf{Additional visual comparison on DAVIS} (same layout as \cref{fig:davis_qual}).
  The camera zooms out from the swing (rows 1 and 2) and rises by $60^\circ$ over the display shelf, the skate bowl and the railway crossing (rows 3 to 8).
  {\ourmethod} shows the whole swing frame after the zoom-out and looks down on the other three scenes.
  DepthDirector~\cite{depthdirector} breaks apart on the swing (rows 1 and 2) and the crossing (rows 7 and 8) and turns toward the sky above the skate bowl (rows 5 and 6).
  Wan-Fun-Cam~\cite{wan} moves closer to the swing instead of away from it (rows 1 and 2), stays at the height of the input at the shelf (rows 3 and 4) and turns into noise on the skate bowl and the crossing (rows 6 and 8).}
  \label{fig:supp_davis_more}
\end{figure*}

\vspace{1mm}
\noindent\textbf{DAVIS.}
\Cref{fig:supp_davis_more} shows four more DAVIS~\cite{davis_2016, davis_2017} videos with all methods of \cref{fig:davis_qual}.

\vspace{1mm}
\noindent\textbf{Ablation study: fine-tuning strategy.}
\Cref{tab:ablation_ft} compares fine-tuning the self-attention layers and the patch embedding, as in \cref{sec:method_model}, with LoRA~\cite{lora} of rank 32 on the attention and feed-forward layers on DAVIS and CO3D.
Fine-tuning the attention layers gives better visual quality on both datasets, \eg, FID~\cite{fid} 41.26 vs 44.12 on DAVIS and 56.0 vs 65.4 on CO3D, and a lower rotation error on DAVIS.
LoRA gives slightly lower translation error and CamMC on DAVIS and a higher anchor IoU on CO3D.
\begin{table*}[t]
  \centering
  \caption{\textbf{Attention fine-tuning versus LoRA.} LoRA~\cite{lora} of rank 32 on the attention and feed-forward layers, trained with a learning rate of $10^{-4}$, versus fine-tuning the self-attention layers and the patch embedding with a learning rate of $5\times10^{-5}$. Both are trained on the same data for 10K steps with a batch size of 32. The attention fine-tuning row is {\ourmethod}; the cases and metrics are those of its DAVIS and CO3D comparisons (\cref{tab:davis_bidi,tab:co3d}). Better in \textbf{bold}.}
  \label{tab:ablation_ft}
  \footnotesize
  \setlength{\tabcolsep}{2.2pt}
  \begin{tabular*}{\textwidth}{@{\extracolsep{\fill}}lccccccccccccc}
    \toprule
    & \multicolumn{7}{c}{DAVIS} & \multicolumn{6}{c}{CO3D} \\
    \cmidrule(r){2-8} \cmidrule(r){9-14}
    Method & Rot.$\downarrow$ & Trans.$\downarrow$ & CamMC$\downarrow$ & FID$\downarrow$ & VBench$\uparrow$ & CLIP-V$\uparrow$ & CLIP-F$\uparrow$ & PSNR$\uparrow$ & SSIM$\uparrow$ & LPIPS$\downarrow$ & FID$\downarrow$ & IoU$\uparrow$ & CLIP-V$\uparrow$ \\
    \midrule
    LoRA ($r{=}32$) & 4.65 & \textbf{0.026} & \textbf{0.154} & 44.12 & 82.34 & 87.89 & 97.10 & 13.77 & 0.503 & 0.517 & 65.4 & \textbf{0.829} & 85.23 \\
    Attention FT ({\ourmethod}) & \textbf{4.60} & 0.028 & 0.159 & \textbf{41.26} & \textbf{82.89} & \textbf{88.53} & \textbf{97.34} & \textbf{14.08} & \textbf{0.530} & \textbf{0.500} & \textbf{56.0} & 0.814 & \textbf{86.59} \\
    \bottomrule
  \end{tabular*}
\end{table*}

\begin{figure*}[t]
  \centering
  \includegraphics[width=\linewidth]{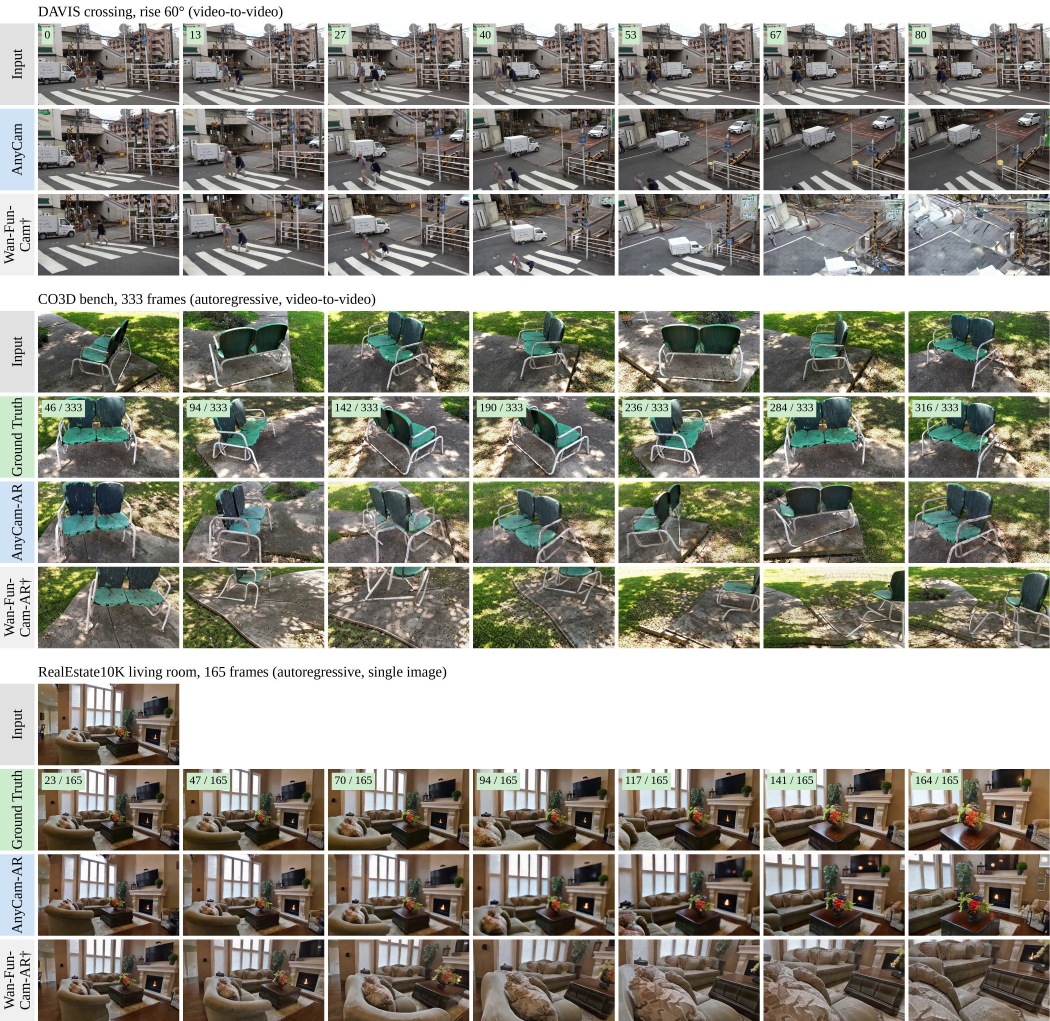}
  \caption{\textbf{Comparison with the Plücker-ray models.}
  Wan-Fun-Cam$^\dagger$ and Wan-Fun-Cam-AR$^\dagger$ are trained on our data with our recipe and receive the camera as Plücker rays instead of the anchor rendering.
  Top: DAVIS~\cite{davis_2016, davis_2017} with a $60^\circ$ rise, {\ourmethod} and Wan-Fun-Cam$^\dagger$; the numbers on the input frames give the frame index.
  Middle: a 333-frame CO3D~\cite{co3d} video (166 frames out and back) generated from an 81-frame source video; bottom: a 165-frame RealEstate10K~\cite{re10k} video generated from a single image; {\ourmethodar} and Wan-Fun-Cam-AR$^\dagger$; the labels on the ground-truth frames give the frame index and the video length.
  Wan-Fun-Cam$^\dagger$ distorts the street in the last frames of the rise.
  Wan-Fun-Cam-AR$^\dagger$ drifts off the bench, which the frame border cuts off from frame 94 on.
  On RealEstate10K, it moves into the armchair instead of toward the coffee table.
  {\ourmethod} and {\ourmethodar} follow the target views.}
  \label{fig:supp_plucker}
\end{figure*}

\vspace{1mm}
\noindent\textbf{Ablation study: camera representation.}
\Cref{fig:supp_plucker} compares {\ourmethod} with Wan-Fun-Cam$^\dagger$ and {\ourmethodar} with Wan-Fun-Cam-AR$^\dagger$.
The comparison isolates the camera representation: both pairs share the training data and the training recipe (\cref{sec:supp_train}); Wan-Fun-Cam$^\dagger$ and Wan-Fun-Cam-AR$^\dagger$ receive the camera as Plücker rays instead of the anchor rendering.
Due to the scale ambiguity of the Plücker-ray representation, Wan-Fun-Cam$^\dagger$ distorts the scene under a large rise, and Wan-Fun-Cam-AR$^\dagger$ drifts away from the target views over long videos.

\vspace{1mm}
\noindent\textbf{Long videos.}
\Cref{tab:co3d_long_full} reports the long CO3D videos on all frames, on the unseen views and on the way back.
\begin{table*}[t]
      \centering
      \caption{\textbf{Long videos on CO3D, all metrics.}
      30 sequences; the source is an 81-frame clip played forward and backward to the target length, and the target follows the real CO3D cameras around the object for 165 or 249 frames, or for 166 frames and back (333 frames).
      PSNR, SSIM and LPIPS compare each generated frame with the ground-truth frame of the same camera, and oPSNR, oSSIM and oLPIPS do so inside the object mask.
      Unseen views are the ground-truth frames whose camera is at least $45^\circ$ in azimuth away from the arc covered by the source clip, \ie sides of the object the input never shows; the way back is the second half of the 333-frame videos.
      Anchor IoU compares the anchor tracked in the generated video with the object mask, and Lost is the fraction of frames where the track is lost.
      Best in \textbf{bold}, second \underline{underlined}, ranked within each group.}
      \label{tab:co3d_long_full}
      \footnotesize
      \setlength{\tabcolsep}{2.2pt}
      \begin{tabular*}{\textwidth}{@{\extracolsep{\fill}}lcccccccc}
    \toprule
    & \multicolumn{6}{c}{Ground Truth} & \multicolumn{2}{c}{Anchor} \\
    \cmidrule(r){2-7} \cmidrule(r){8-9}
    Method & PSNR$\uparrow$ & SSIM$\uparrow$ & LPIPS$\downarrow$ & oPSNR$\uparrow$ & oSSIM$\uparrow$ & oLPIPS$\downarrow$ & Anchor IoU$\uparrow$ & Lost$\downarrow$ \\
    \midrule
    \multicolumn{9}{l}{\textit{165 frames, all frames}} \\
    \midrule
    HY-WorldPlay & 11.23 & 0.371 & 0.689 & 9.51 & 0.271 & 0.621 & 0.195 & 0.077 \\
    InSpatio-World & 11.98 & 0.448 & 0.634 & \underline{10.74} & \underline{0.345} & \underline{0.570} & \underline{0.243} & 0.094 \\
    Wan-Fun-Cam-AR$^\dagger$ & \underline{12.70} & \underline{0.464} & \underline{0.631} & 10.72 & 0.328 & 0.594 & 0.207 & \underline{0.017} \\
    {\ourmethodar} & \textbf{14.34} & \textbf{0.516} & \textbf{0.506} & \textbf{14.37} & \textbf{0.429} & \textbf{0.366} & \textbf{0.782} & \textbf{0.001} \\
    \midrule
    \multicolumn{9}{l}{\textit{165 frames, unseen views}} \\
    \midrule
    HY-WorldPlay & 10.92 & 0.364 & 0.708 & 9.16 & 0.265 & 0.641 & 0.158 & 0.051 \\
    InSpatio-World & 11.56 & 0.440 & 0.657 & 10.37 & \underline{0.339} & \underline{0.587} & \underline{0.200} & 0.108 \\
    Wan-Fun-Cam-AR$^\dagger$ & \underline{12.48} & \underline{0.459} & \underline{0.646} & \underline{10.46} & 0.325 & 0.607 & 0.157 & \underline{0.018} \\
    {\ourmethodar} & \textbf{13.99} & \textbf{0.512} & \textbf{0.523} & \textbf{14.11} & \textbf{0.420} & \textbf{0.380} & \textbf{0.771} & \textbf{0.002} \\
    \midrule
    \multicolumn{9}{l}{\textit{249 frames, all frames}} \\
    \midrule
    HY-WorldPlay & 11.08 & 0.344 & 0.699 & 9.50 & 0.257 & 0.630 & 0.147 & 0.197 \\
    InSpatio-World & 12.00 & 0.445 & 0.637 & 10.73 & \underline{0.344} & \underline{0.569} & \underline{0.243} & 0.106 \\
    Wan-Fun-Cam-AR$^\dagger$ & \underline{12.76} & \underline{0.457} & \underline{0.631} & \underline{10.90} & 0.331 & 0.580 & 0.226 & \underline{0.024} \\
    {\ourmethodar} & \textbf{14.34} & \textbf{0.509} & \textbf{0.509} & \textbf{14.17} & \textbf{0.421} & \textbf{0.381} & \textbf{0.764} & \textbf{0.000} \\
    \midrule
    \multicolumn{9}{l}{\textit{249 frames, unseen views}} \\
    \midrule
    HY-WorldPlay & 10.77 & 0.336 & 0.719 & 9.14 & 0.251 & 0.650 & 0.102 & 0.177 \\
    InSpatio-World & 11.62 & 0.435 & 0.661 & 10.34 & \underline{0.337} & \underline{0.588} & \underline{0.196} & 0.114 \\
    Wan-Fun-Cam-AR$^\dagger$ & \underline{12.56} & \underline{0.453} & \underline{0.646} & \underline{10.61} & 0.327 & 0.592 & 0.185 & \underline{0.022} \\
    {\ourmethodar} & \textbf{14.05} & \textbf{0.505} & \textbf{0.529} & \textbf{13.85} & \textbf{0.412} & \textbf{0.401} & \textbf{0.750} & \textbf{0.000} \\
    \midrule
    \multicolumn{9}{l}{\textit{333 frames, all frames}} \\
    \midrule
    HY-WorldPlay & 10.82 & 0.334 & 0.719 & 9.59 & 0.250 & 0.635 & 0.115 & 0.250 \\
    InSpatio-World & 11.74 & 0.441 & \underline{0.649} & 10.57 & \underline{0.332} & \underline{0.585} & \underline{0.239} & 0.074 \\
    Wan-Fun-Cam-AR$^\dagger$ & \underline{12.55} & \underline{0.451} & \underline{0.649} & \underline{10.74} & 0.324 & 0.597 & 0.175 & \underline{0.026} \\
    {\ourmethodar} & \textbf{14.06} & \textbf{0.511} & \textbf{0.526} & \textbf{14.00} & \textbf{0.417} & \textbf{0.386} & \textbf{0.746} & \textbf{0.000} \\
    \midrule
    \multicolumn{9}{l}{\textit{333 frames, unseen views}} \\
    \midrule
    HY-WorldPlay & 10.66 & 0.330 & 0.729 & 9.42 & 0.245 & 0.646 & 0.092 & 0.217 \\
    InSpatio-World & 11.40 & 0.434 & 0.668 & 10.22 & \underline{0.326} & \underline{0.602} & \underline{0.198} & 0.085 \\
    Wan-Fun-Cam-AR$^\dagger$ & \underline{12.48} & \underline{0.450} & \underline{0.655} & \underline{10.64} & 0.323 & 0.604 & 0.159 & \underline{0.025} \\
    {\ourmethodar} & \textbf{13.88} & \textbf{0.509} & \textbf{0.536} & \textbf{13.86} & \textbf{0.412} & \textbf{0.394} & \textbf{0.738} & \textbf{0.001} \\
    \midrule
    \multicolumn{9}{l}{\textit{333 frames, way back}} \\
    \midrule
    HY-WorldPlay & 10.33 & 0.294 & 0.756 & 9.47 & 0.222 & 0.657 & 0.040 & 0.396 \\
    InSpatio-World & 11.45 & 0.433 & \underline{0.665} & 10.39 & \underline{0.323} & \underline{0.592} & \underline{0.232} & 0.080 \\
    Wan-Fun-Cam-AR$^\dagger$ & \underline{12.40} & \underline{0.439} & \underline{0.665} & \underline{10.75} & 0.320 & 0.602 & 0.143 & \underline{0.039} \\
    {\ourmethodar} & \textbf{13.73} & \textbf{0.505} & \textbf{0.548} & \textbf{13.59} & \textbf{0.402} & \textbf{0.412} & \textbf{0.718} & \textbf{0.000} \\
    \bottomrule
  \end{tabular*}
\end{table*}

\section{Discussion}
\label{sec:supp_discussion}

\begin{figure*}[t]
  \centering
  \includegraphics[width=\linewidth]{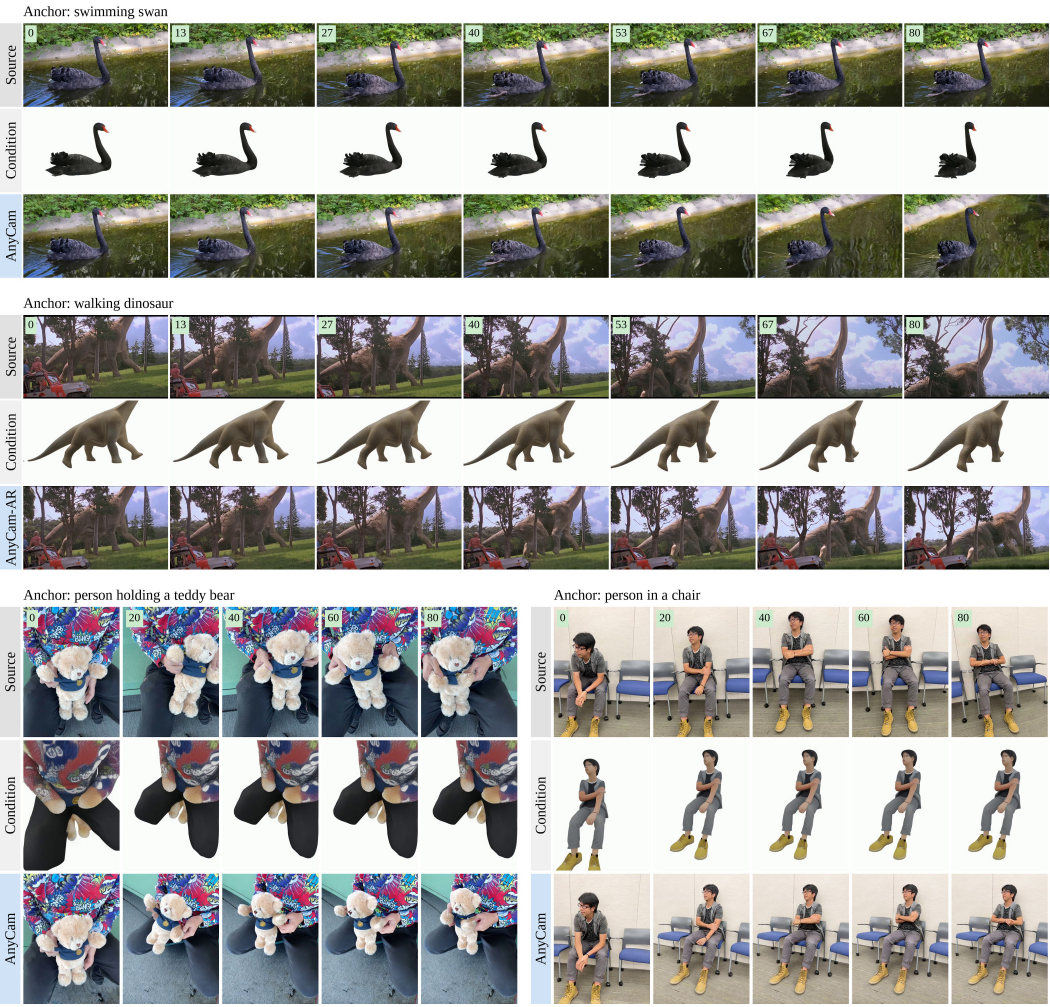}
  \caption{\textbf{Moving and deforming anchors.}
  The anchor need not be static or rigid: a swimming swan (DAVIS~\cite{davis_2016, davis_2017}, row 1), a walking dinosaur (row 2), and a person from the DyCheck dataset~\cite{dycheck} (bottom left: holding a teddy bear; bottom right: sitting in a chair).
  The condition renders the anchor reconstructed from one frame, while the generated video keeps the motion of the source, \eg, the swimming swan, the walking legs, and the moving hands and arms.
  Row 2 shows {\ourmethodar}; all other examples show {\ourmethod}.}
  \label{fig:supp_anchor_cases}
\end{figure*}
\begin{figure*}[t]
  \centering
  \includegraphics[width=\linewidth]{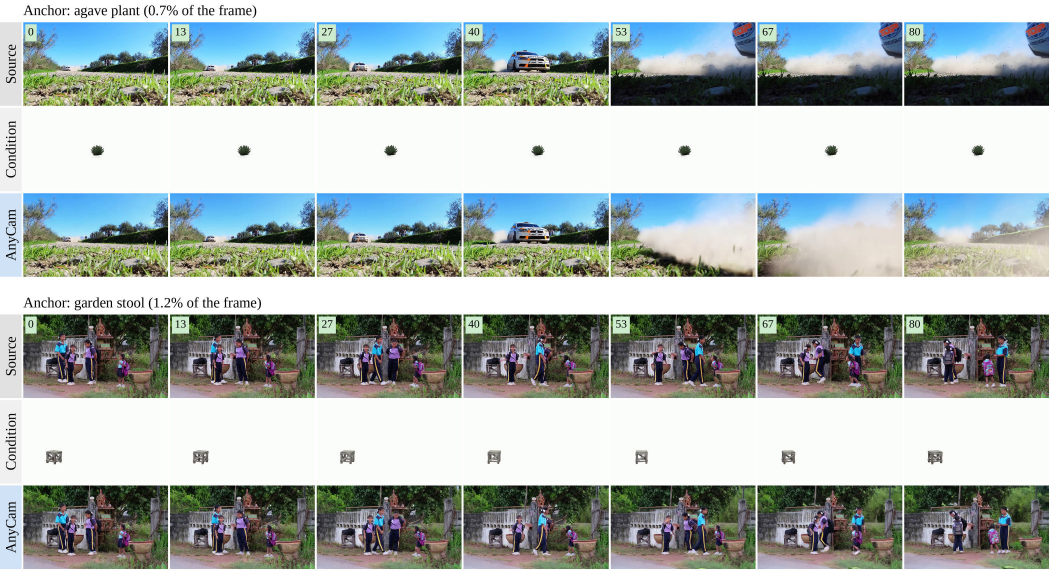}
  \caption{\textbf{Failure case: small anchors.}
  When the anchor covers only about 1\% of the frame, its rendering gives little information about the rotation of the camera.
  For the agave plant (row 1, 0.7\% of the frame), the output camera barely moves, and for the garden stool (row 2, 1.2\%) it turns much less than the target orbit.
  Both rows show {\ourmethod}.}
  \label{fig:supp_anchor_cases_b}
\end{figure*}

\begin{figure*}[p]   %
  \centering
  \includegraphics[width=\linewidth]{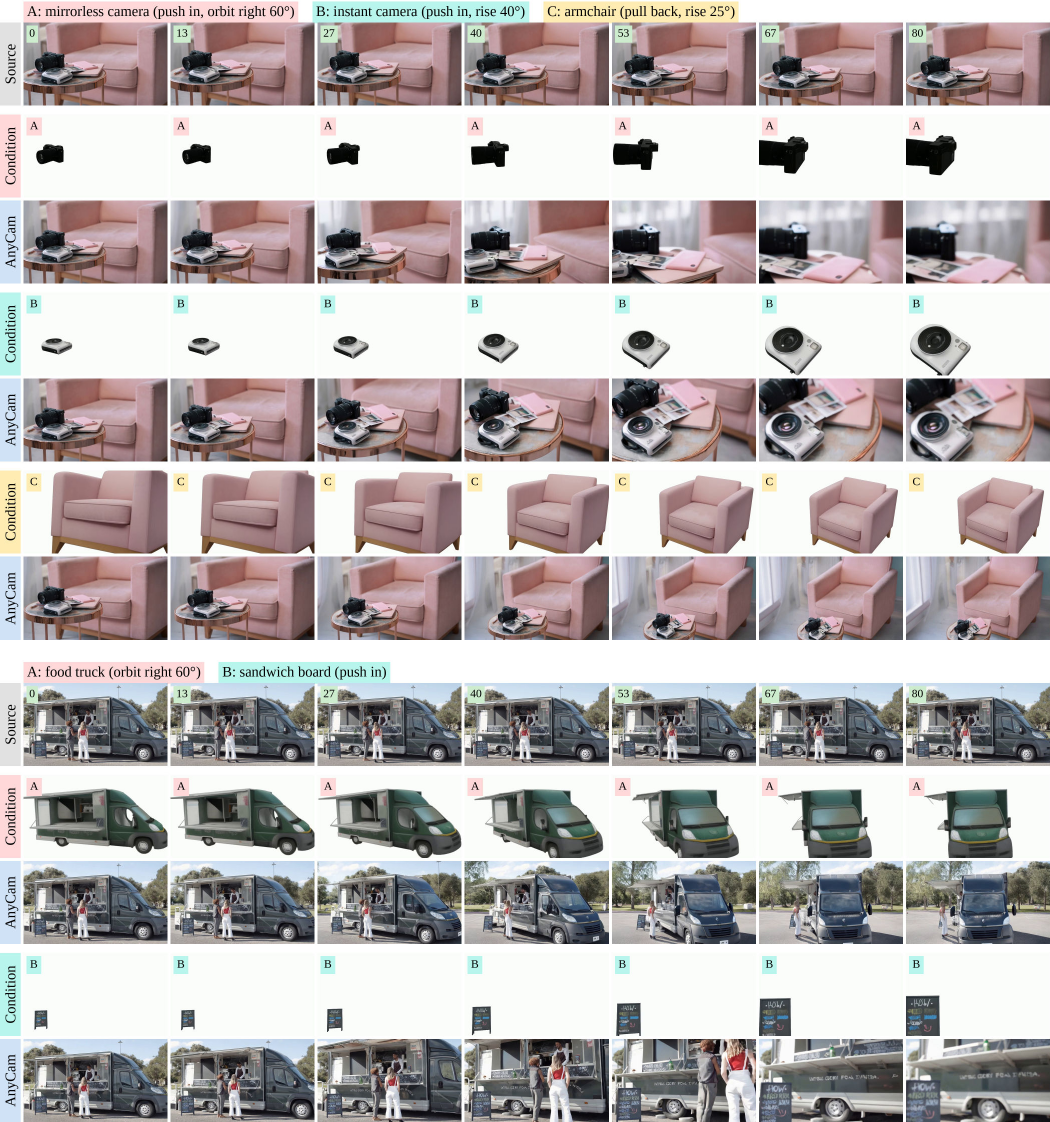}
  \caption{\textbf{Different anchors in the same video.}
  For each source video, the user picks two or three objects as anchors and a target trajectory around each of them. The tags above each example name the anchors and their trajectories.
  Each anchor has its own trajectory, so the rows show camera control through different objects of one video, not one camera applied to several anchors.
  Each condition renders only its anchor. The numbers on the source frames give the frame index. All outputs are from {\ourmethod}.
  In the first video, the camera pushes in toward the mirrorless camera or the instant camera on the table, or pulls back and rises over the armchair.
  In the second video, it orbits around the food truck toward its front or pushes in to the sandwich board.}
  \label{fig:supp_anchors_multi}
\end{figure*}
\begin{figure*}[p]   %
  \centering
  \includegraphics[width=\linewidth]{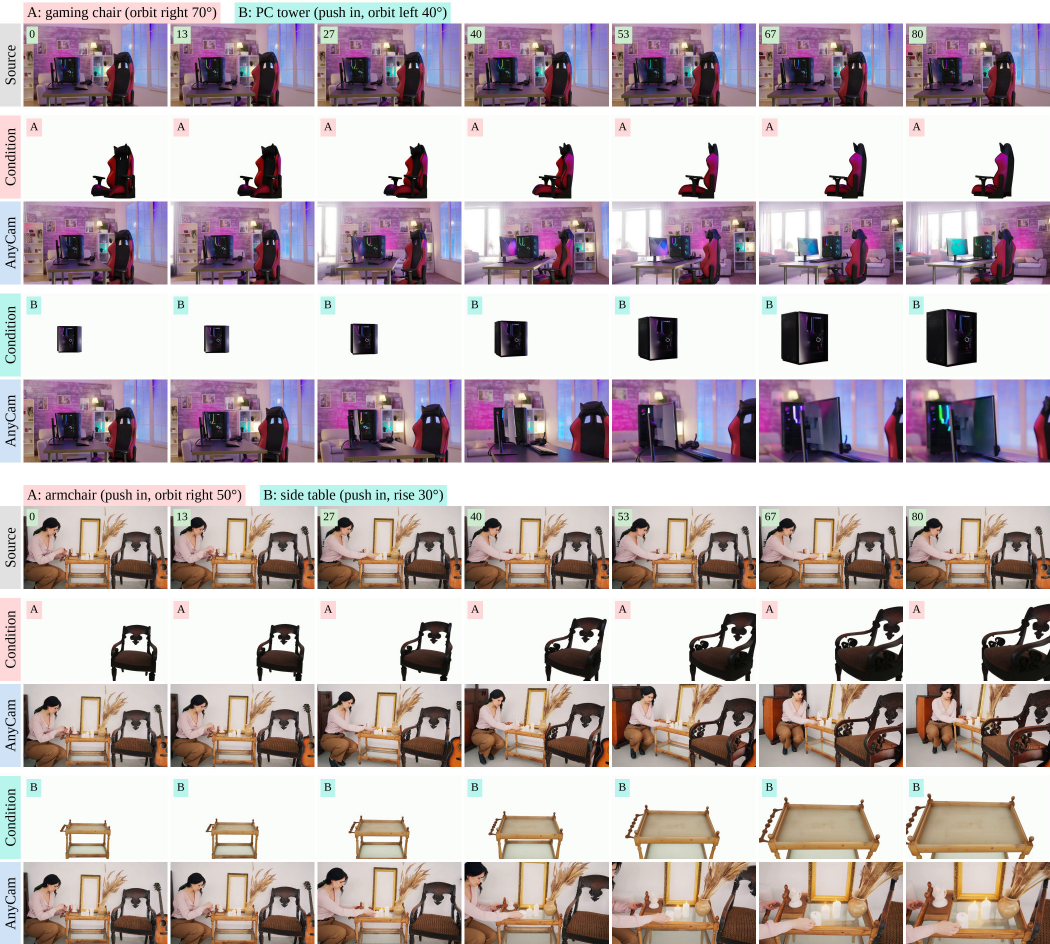}
  \caption{\textbf{Different anchors in the same video (continued).} Same setting as \cref{fig:supp_anchors_multi}. All outputs are from {\ourmethod}.
  In the gaming room, the output orbits around the gaming chair or moves toward the PC tower. In the second video, it orbits around the carved armchair or rises over the side table.}
  \label{fig:supp_anchors_multi_b}
\end{figure*}
\begin{figure*}[p]   %
  \centering
  \includegraphics[width=\linewidth]{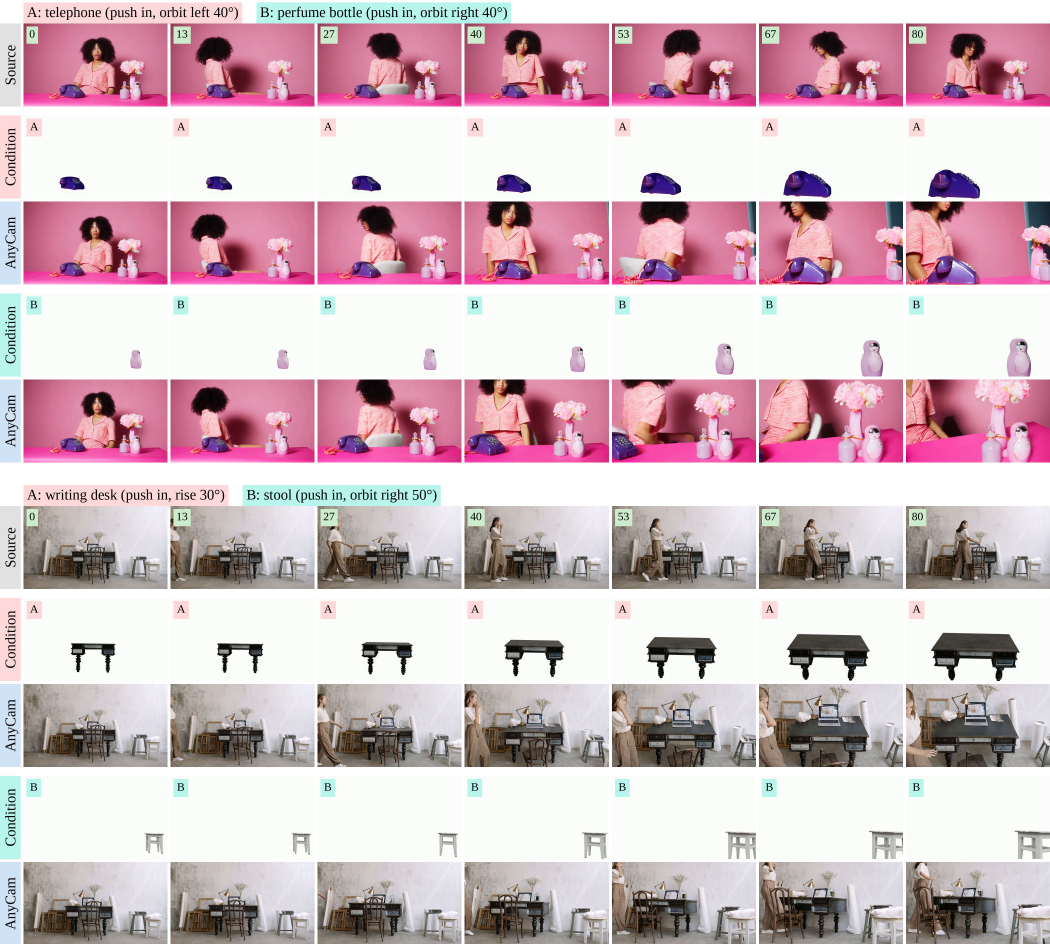}
  \caption{\textbf{Different anchors in the same video (continued).} Same setting as \cref{fig:supp_anchors_multi}. All outputs are from {\ourmethod}.
  The output moves toward the telephone or the perfume bottle while the person keeps moving as in the source. In the studio, it rises over the writing desk or turns toward the stool.}
  \label{fig:supp_anchors_multi_c}
\end{figure*}

\begin{figure*}[p]   %
  \centering
  \includegraphics[width=\linewidth]{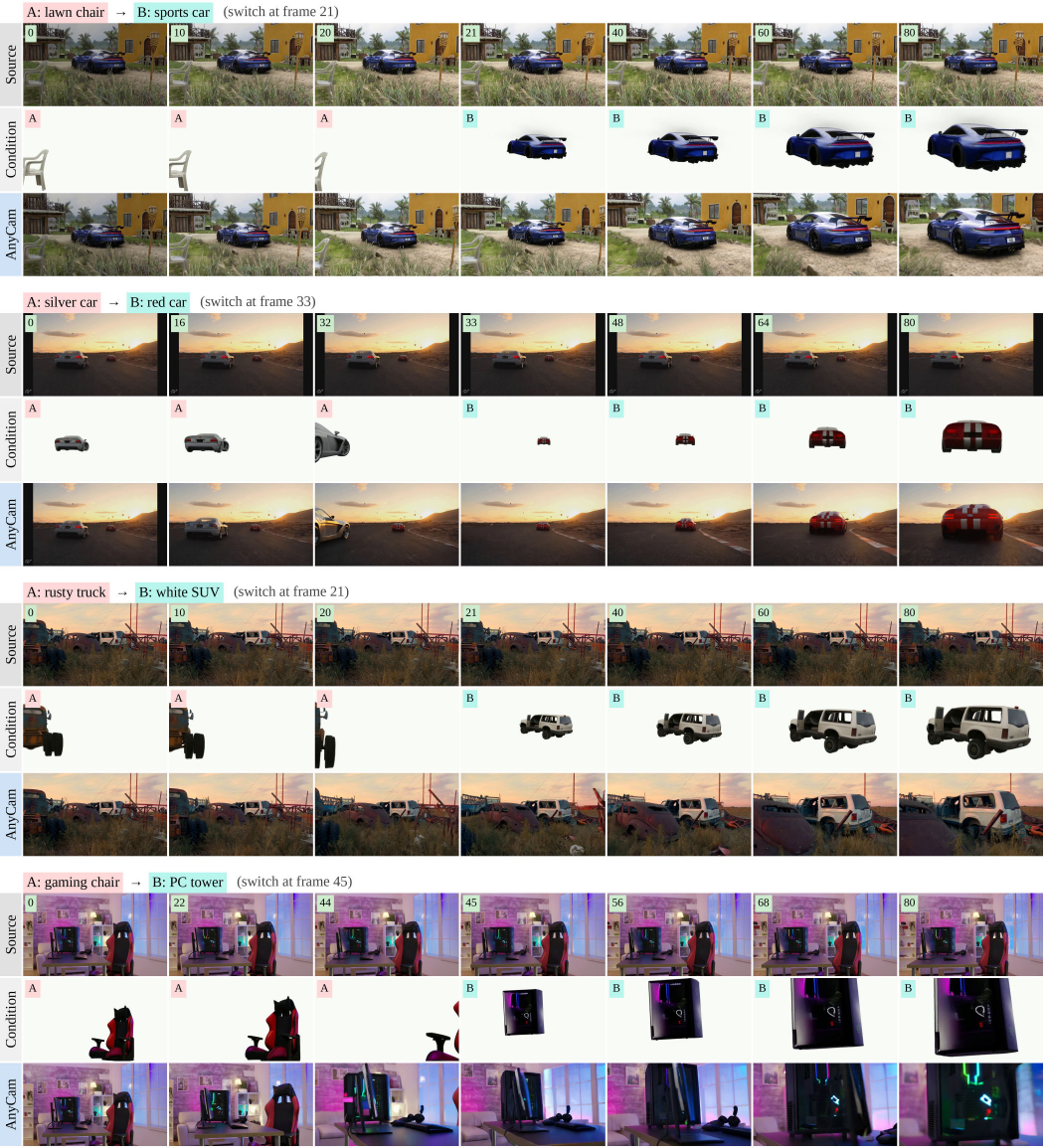}
  \caption{\textbf{The anchor leaves the frame.}
  The camera moves past anchor A until A leaves the view; starting at the switch frame, the condition renders anchor B, a second object reconstructed from the first source frame, and the camera continues toward it.
  The numbers on the source frames give the frame index; tags A and B mark the anchor in the condition.
  All rows show {\ourmethod}. The output keeps moving after A has left and approaches anchor B (the sports car, the red car, the white SUV and the PC tower in rows 1 to 4).}
  \label{fig:supp_e3_leave}
\end{figure*}

\begin{figure*}[p]   %
  \centering
  \includegraphics[width=\linewidth]{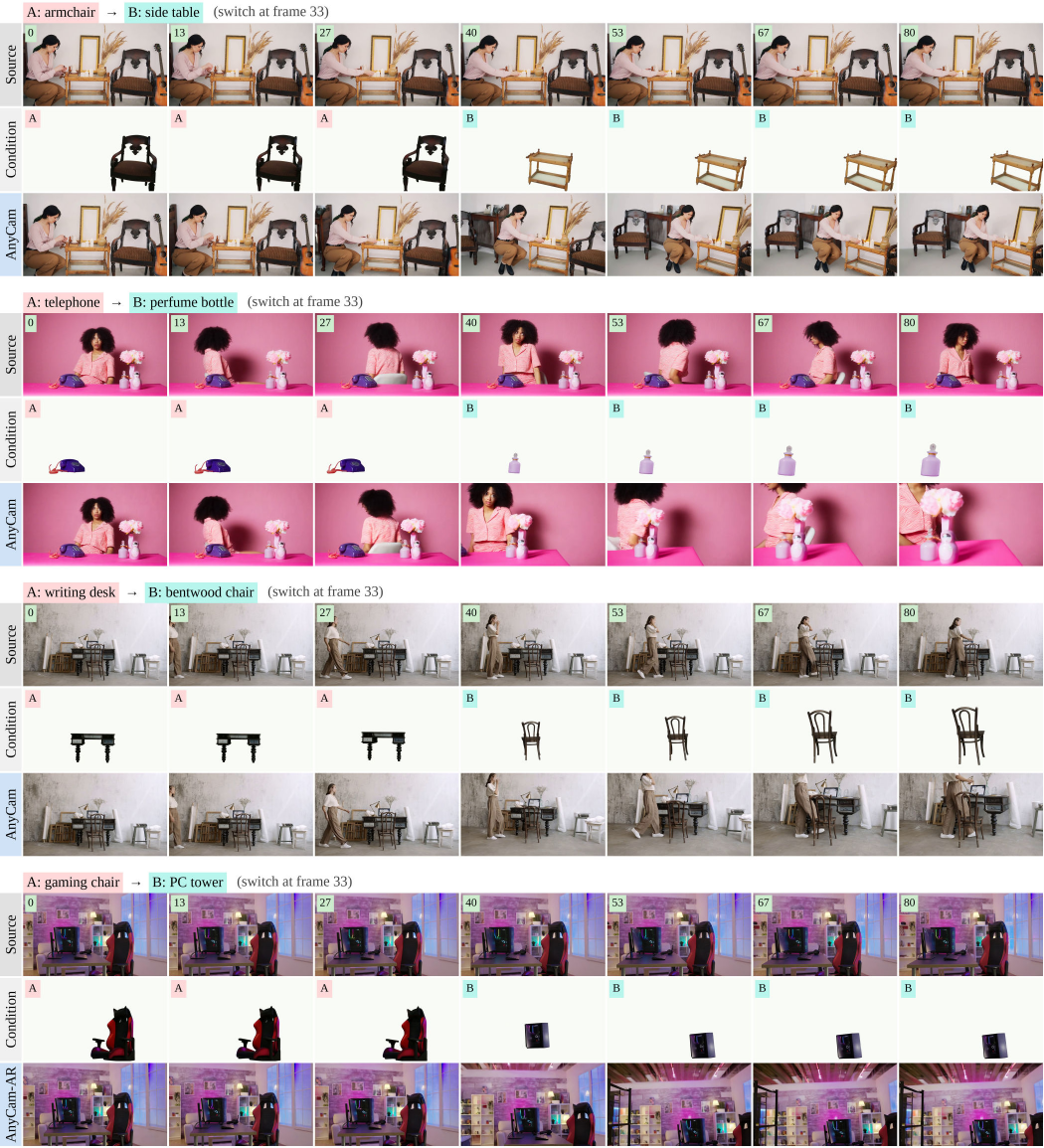}
  \caption{\textbf{Switching the anchor during a video.}
  The target camera follows one continuous trajectory. The condition renders anchor A up to frame 32 and anchor B from frame 33 (tags A and B), and both anchors are reconstructed from the first source frame.
  The numbers on the source frames give the frame index.
  Rows 1 to 3 show {\ourmethod}, row 4 shows {\ourmethodar}.
  After the switch, the output moves toward the side table, the perfume bottle, the bentwood chair and the PC tower.}
  \label{fig:supp_e1_switch}
\end{figure*}

\noindent\textbf{Which objects can be anchors.}
Although our automatic anchor selection favors rigid and stationary objects for reliable training data construction, the method itself does not impose these restrictions.
The anchor only has to be visible in the source video and reconstructable from one frame.
It need not be static or rigid: \cref{fig:supp_anchor_cases} shows a swimming swan, a walking dinosaur and a person as anchors.
The condition renders the anchor reconstructed from a single frame, while the generated video keeps the motion of the source, \eg, the moving legs and arms, so the rendering acts as a camera signal rather than as the content of the output.
One video can thus be controlled through different objects: \cref{fig:supp_anchors_multi,fig:supp_anchors_multi_b,fig:supp_anchors_multi_c} show two or three anchors per video, each with its own camera path.

\vspace{1mm}
\noindent\textbf{Switching the anchor.}
The anchor may leave the view as the camera moves, and then we can no longer control the camera through it.
To keep control of the camera, we can switch to another object that is still visible.
All anchors of a video are reconstructed from its first frame and placed in the same camera frame, so the condition can change from anchor A to anchor B while the target camera follows one continuous trajectory.
In \cref{fig:supp_e3_leave}, the camera moves past anchor A, and after the switch to anchor B it keeps moving along the trajectory.
The anchor can also be switched while it is still in view: in \cref{fig:supp_e1_switch}, the condition renders anchor A up to frame 32 and anchor B from frame 33, and after the switch the output moves toward B.

\vspace{1mm}
\noindent\textbf{Failure cases and limitations.}
The rendering conveys rotation through the appearance of the anchor, so a small anchor or a rotationally symmetric one gives weak control of the rotation; for anchors covering about 1\% of the frame, the output camera turns much less than the target (\cref{fig:supp_anchor_cases_b}).
Scenes without a distinct object, such as open landscapes, offer no suitable anchor.
A poor reconstruction still conveys the camera (\cref{fig:sam3d_condition}), but a reconstruction in the wrong orientation or scale gives a wrong camera.

\vspace{1mm}
\noindent\textbf{Societal impact.}
Like other video generation methods, {\ourmethod} can re-render real videos from viewpoints that were never captured, which could be used to create misleading content; generated videos should be marked as such.

\section{Method and Training Details}
\label{sec:supp_impl}

\subsection{Anchor Selection and Reconstruction}
\label{sec:supp_anchor}
\noindent\textbf{Vision-language model.}
We use Qwen3-VL-32B-Instruct~\cite{qwen3vl} with greedy decoding.
It sees 8 frames sampled uniformly from the video and returns the anchor as a short noun phrase together with a JSON record of which criteria the object satisfies.
The prompt is shown in \cref{fig:supp_prompt}.
It asks for an object that is visible in all frames, large in at least one frame, prominent, rigid and stationary, compact, clearly oriented, and describable by a short noun phrase, and it lists categories to prefer and to reject.
The same prompt selects the anchors of the evaluation sets (\cref{sec:exp_setup}).
These criteria give clean anchors for building the training data; at inference, the user can pick any object, and \cref{sec:supp_discussion} discusses which anchors work well and which do not.
For MultiCamVideo~\cite{recammaster}, the model looks at the camera with the largest translation.

\vspace{1mm}
\noindent\textbf{Segmentation.}
SAM~3~\cite{sam3} finds the anchor from its phrase in one frame.
SAM~2~\cite{sam2} then propagates the mask forward and backward from this frame.

\vspace{1mm}
\noindent\textbf{Reconstruction.}
The reconstruction frame is the frame in which the anchor mask is largest and not cut by the image border.
We run SAM~3D Objects~\cite{sam3d} with a fixed seed and with its layout post-processing enabled, which noticeably improves the predicted pose.

\begin{figure*}[t]
\centering
\begin{minipage}{0.97\linewidth}
\scriptsize
\begin{verbatim}
You select an anchor object for a downstream 3D reconstruction pipeline. The object you pick will be (a) tracked
   across the video by an open-vocabulary segmentation model and (b) reconstructed as a 3D mesh from a single frame.
   You will see 8 frames sampled uniformly from a video.

Pick ONE physical object that satisfies ALL of the following:
  1. Visible in ALL 8 frames (never leaves the view).
  2. LARGE in the frame in at least one of the 8 shown frames - the object should occupy AT LEAST ~20% of the image
     area (roughly one fifth of the picture) in the frame where it is largest. A small, distant, or thumbnail-sized
     object is NOT acceptable, even if it is central. When in doubt, pick a closer / larger object.
  3. Prominent and relatively central in the scene, not a background detail.
  4. Rigid and stationary - NOT a person, animal, plant, water, cloth, curtain, flag, or anything that
     moves or deforms.
  5. A discrete, self-contained physical object with clear, compact 3D volume - something a human could point at and
     call 'that object'.
  6. HAS A CLEAR ORIENTATION - the object must look visually DIFFERENT from different sides (front vs back vs left vs
     right). This lets a downstream model infer camera rotation from how the object appears. Objects with rotational
     or near-rotational symmetry (round vases, bowls, baskets, balls, drums, barrels, cylindrical pots, plates,
     unlabeled cans) are NOT acceptable because rotating the camera around them barely changes their appearance.
     Prefer objects with a clear 'front' - a chair back, a sofa's L-shape, a TV screen, a car's front, a statue's
     face, an appliance's control panel, a labeled box.
  7. Describable as a short, concrete noun phrase (e.g., 'wooden dining chair', 'gray sofa', 'stainless steel
     refrigerator', 'red toaster').

PREFER (good 3D reconstruction candidates):
  - Furniture: chair, sofa, table, bed, cabinet, shelf, stool, ottoman, desk
  - Decor / small objects: vase, sculpture, statue, lamp, basket, box, clock, book, mug, bowl, bottle, toy,
     picture frame
  - Appliances / electronics: TV, microwave, refrigerator, computer, speaker
  - Vehicles: car, bicycle, motorcycle
  - Musical instruments: piano, guitar, drum

REJECT (bad for the pipeline - set object=null if these are the only options):
  - Architectural / building elements: wall, floor, ceiling, staircase, archway, doorway, column, pillar, window
     frame, fireplace built into a wall
  - Thin / repetitive structures: fence, railing, pipe, wire, ladder, cable
  - Flat surfaces without volume: rug, mat, painting hanging flat on a wall, poster, mirror
  - Transparent / reflective items: glass table, glass door, mirror, water
  - Extremely complex geometry: chandelier with many arms, tree, bush, leafy plant, curtain, drapery
  - Fixtures embedded in the scene: kitchen counter, built-in shelf, sink attached to a vanity
  - Rotationally symmetric / no clear orientation: round vase, bowl, basket, ball, drum, barrel, cylindrical pot,
     plate, unlabeled can, single wheel, sphere, disc - anything that looks nearly identical from all sides

Reply with STRICTLY this JSON and nothing else:
{
  "object": "<short noun phrase>" or null,
  "reason": "<one short sentence>",
  "visible_in_all_frames": true|false,
  "central": true|false,
  "large_in_frame": true|false,
  "distinguishable_orientation": true|false,
  "reconstructable_3d": true|false
}
Set "object" to null if no candidate satisfies ALL SEVEN criteria. Prefer null over a bad choice.
\end{verbatim}
\end{minipage}
\caption{\textbf{Anchor selection prompt.} System prompt given to Qwen3-VL-32B together with 8 frames of the video.}  %
\label{fig:supp_prompt}
\end{figure*}

\subsection{Anchor Alignment}
\label{sec:supp_align}
\noindent\textbf{View selection.}
For each frame, we compute the unit direction from the camera to the object in the anchor frame and then select views by farthest-point sampling on these directions, starting from the reconstruction frame.
Sampling directions rather than camera positions makes the selected views cover different sides of the object even when the camera moves mostly along one path.

\vspace{1mm}
\noindent\textbf{Stage 1: scale.}
We evaluate 20 scales log-spaced in $[0.05, 20]$ and refine the best one by 10 iterations of golden-section search between its two neighbors.
The objective is the mean IoU over 20 selected views between the rendered object ($\alpha > 0.2$) and the SAM~2 mask.
We use IoU in this stage because a loss that only penalizes rendered pixels outside the mask is minimized by an empty rendering.

\vspace{1mm}
\noindent\textbf{Stage 2: similarity transform.}
Starting from the scale of stage 1, we optimize a rotation (quaternion), a translation and a log-scale applied to $\mathcal{G}$ about its centroid, minimizing
\begin{equation}
  \mathcal{L}_{\text{align}} = \frac{1}{|\mathcal{V}|\,|\Omega|}\sum_{i\in\mathcal{V}}\sum_{\mathbf{p}\in\Omega}\big( (g * \alpha_i)(\mathbf{p}) - (g * M_i)(\mathbf{p}) \big)^2,
  \label{eq:align}
\end{equation}
where $\mathcal{V}$ is the set of 20 selected views, $\Omega$ the set of pixels, $\alpha_i \in [0, 1]$ the rendered opacity of view $i$, used without thresholding, $M_i$ its SAM mask and $g$ a separable Gaussian filter with a $15\times15$ kernel and a standard deviation of 4 pixels, applied to both, which gives a smooth gradient when the rendered object and the mask are slightly misaligned.
We run Adam for 250 iterations with learning rates $2\times10^{-4}$, $10^{-3}$ and $5\times10^{-4}$ for rotation, translation and log-scale, add an $L_2$ penalty of weight $0.01$ on each of the three updates, and clip the gradient norm at $0.5$, since gradients through the rasterized object mask are noisy.
Every 20 iterations we compute the IoU ($\alpha > 0.5$) and keep the best state.
All renderings use the intrinsics of the dataset.

\vspace{1mm}
\noindent\textbf{MultiCamVideo.}
The 10 cameras of a MultiCamVideo~\cite{recammaster} scene start from the same pose and share one metric world frame, so each scene is reconstructed and aligned once and rendered for all cameras.
The alignment uses the masks of all 10 cameras.
Since the world frame of MultiCamVideo is gravity-aligned, we also fit an upright version, in which the anchor's axis closest to the world up direction is rotated onto it, and use this fit when it matches the masks about as well.

\subsection{Training Data}
\label{sec:supp_data}

\noindent\textbf{Overview.}
The training dataset contains 7{,}810 pairs from 13{,}600 MultiCamVideo~\cite{recammaster} scenes and 5{,}217 pairs from 9{,}894 DL3DV~\cite{dl3dv} videos.
Every sample consists of the source video, the source condition (SAM~2 cut-out of the anchor on white), the target condition (rendering of the aligned anchor on white) and the target video, all with 81 frames at $832\times480$.

\vspace{1mm}
\noindent\textbf{MultiCamVideo pairs.}
We build one source-target camera pair per scene.
The source camera must contain a frame that shows the whole anchor.
The target camera is sampled with a preference for a large change of viewpoint and a large anchor rendering.
Each video is cropped from $1280\times1280$ at a random scale $s \in [0.65, 1]$ to a fixed $832\times480$ window; for the target, the window must keep the anchor rendering visible in every frame.

\vspace{1mm}
\noindent\textbf{DL3DV pairs.}
From each capture we cut two non-overlapping 81-frame windows in which the anchor rendering is visible in every frame.
Among good windows, the second one is chosen to be far from the first in camera position, which favors large viewpoint changes, and the window in which the anchor is more complete serves as the source.
When only one window exists, the source is a single frame repeated 81 times, chosen as a frame in which the anchor is fully inside the image; this gives 926 samples with a static source, which also teaches the model single-image input.  %

\vspace{1mm}
\noindent\textbf{Filtering.}
For MultiCamVideo, a camera is usable only if the anchor is visible in all 81 frames.
We then score every candidate pair with automatic quality features, such as the agreement between the rendering and the SAM~2 masks across cameras and the tilt of the anchor; pairs with high scores are kept, pairs with low scores are removed, and the remaining ones are reviewed manually.
DL3DV~\cite{dl3dv} samples are handled in the same way with a quality score trained on our manual labels.

\subsection{Single-Image Input}
\label{sec:supp_image}
A single input image is repeated to form a static source video.
The anchor is reconstructed from this image.
Stage 1 of the alignment finds the scale from the motion of the source camera, which a single image does not have, so we skip it and run only stage 2 on this single view, with the intrinsics of the dataset, to refine the placement of the anchor in the image.
Instead, the scale is determined from a metric depth estimate: UniDepth-v2~\cite{unidepth} predicts the depth of the input image, which gives the metric scale of the dataset cameras, and the depth of the anchor in the input camera relates this scale to the reconstruction.
All methods get the target trajectory in these metric units.

\subsection{Models and Training}
\label{sec:supp_train}
The bidirectional model fine-tunes the self-attention layers of all DiT~\cite{dit} blocks and the patch embedding (\cref{sec:method_model}) and drops the text prompt with probability $0.1$ during training.
{\ourmethodar} is distilled from the bidirectional model.
A LoRA of rank 128 on this model is trained with teacher forcing~\cite{teacher_forcing} for 3K steps with a learning rate of $10^{-4}$ and a batch size of 32.
DMD~\cite{dmd} then runs for 600 iterations with learning rates of $5\times10^{-6}$ for the generator and $10^{-6}$ for the critic, a batch size of 16, five critic updates per generator update and text guidance of scale 3 for the real score.
For videos longer than 81 frames, each chunk attends to the first chunk and to the 5 most recent chunks, with history noise $\sigma_h=0.1$.

\section{Evaluation Details}
\label{sec:supp_evaldetails}

\subsection{Baselines}
\label{sec:supp_baselines}
All baselines run with their official code and default settings, and their outputs are resized to $832\times480$.
Every method receives the same target trajectory, expressed relative to the first source camera with translations expressed in meters.
Methods that take the source cameras as input receive the same source cameras as {\ourmethod}.
These are the dataset cameras, except on DAVIS~\cite{davis_2016, davis_2017}, which has no camera annotations, where we estimate them with ViPE~\cite{vipe}.
As in their official pipelines, GEN3C~\cite{gen3c} and ReDirector~\cite{redirector} estimate the source geometry with their own ViPE run, InSpatio-World~\cite{inspatio_world} with Depth Anything~3~\cite{da3}, and TrajectoryCrafter~\cite{trajectorycrafter} with its own depth estimate.
HY-WorldPlay~\cite{worldplay} generates from the first source frame and the target trajectory.
3DGS~\cite{gaussian_splatting} is optimized per scene on the 81 source frames with their cameras.
In the single-image setting, the video-to-video methods receive the image as a static video, GEN3C runs in its single-image mode, SEVA~\cite{seva} generates the target views from the image with the sparse-view settings of its official code, and OpenLRM~\cite{lrm}, LGM~\cite{lgm}, TRELLIS~\cite{trellis} and SAM~3D reconstruct the object from the image and render it along the target trajectory.

\subsection{Evaluation Protocol}
\label{sec:supp_eval}
\noindent\textbf{DAVIS.}
The 9 object-centric trajectories on DAVIS~\cite{davis_2016, davis_2017} are adapted from the basic trajectories of ReCamMaster~\cite{recammaster}: arc left and right by $60^\circ$, rise by $60^\circ$, zoom in and out to $0.6$ and $1.5$ times the camera-to-anchor distance, pan left and right by $10^\circ$, and tilt up and down by $5^\circ$.
Since no ground truth is available, camera accuracy is measured on the poses that ViPE~\cite{vipe} estimates from the generated videos, FID compares the generated frames with the source videos, and CLIP-V~\cite{clip} and CLIP-F measure the similarity between the source and the generated frames at the same time step and between consecutive generated frames.

\vspace{1mm}
\noindent\textbf{CO3D.}
The easy, medium and hard targets follow real CO3D~\cite{co3d} camera trajectories at increasing angular distances from the source arc (\cref{sec:supp_results}, \cref{fig:co3d_tiers}).
The single-image set uses the same 102 cases with only the first source frame (53 distinct images); video-to-video methods receive it as a static video.
The long-video set follows the real CO3D cameras around the object for 165 or 249 frames, or for 166 frames and then the same cameras back to source frame 0 (333 frames).

\vspace{1mm}
\noindent\textbf{Mip-NeRF 360 and RealEstate10K.}
On Mip-NeRF 360~\cite{mipnerf360}, each scene has a dense and a sparse trajectory in each direction, yielding four trajectories in total. A dense trajectory reaches a ground-truth photo every 4 frames (20 photos) and a sparse one every 8 frames (10 photos); the dense and the sparse trajectories are optimized separately.
On RealEstate10K~\cite{re10k}, the target starts from the first frame of each clip and follows the real camera path for 165 or 249 frames, or for 208 frames and back (417 frames).

\vspace{1mm}
\noindent\textbf{Metrics.}
Ground-truth metrics compare each generated frame with the ground-truth frame of the same camera, over the full frame and, on CO3D, inside the object mask (oPSNR, oSSIM, oLPIPS~\cite{lpips}).
VBench~\cite{vbench} is $100\times$ the average of subject and background consistency, motion smoothness, aesthetic and imaging quality and temporal flickering.
Rotation and translation errors follow ReDirector~\cite{redirector}: the rotation error is the mean geodesic angle between the relative rotations of all frame pairs of the target and the estimated trajectory, and the translation error is computed by aligning the estimated camera centers to the target ones with a similarity transform and dividing the per-frame distance by the largest distance between two target camera centers.
Anchor IoU tracks the anchor in the generated video with SAM~2, seeded with its mask in the first frame, and compares it with the object mask of the ground truth.
mPSNR, mSSIM and mLPIPS on RealEstate10K are computed only in the part of each ground-truth frame that is visible in the input frame.

\begin{table}[t]
  \centering
  \small
  \caption{\textbf{Streaming cost of {\ourmethodar}} on NVIDIA RTX PRO 6000 Blackwell Max-Q GPUs, median of 4 DAVIS~\cite{davis_2016, davis_2017} videos at $832\times480$.}
  \label{tab:supp_runtime}
  \begin{tabular}{@{}lr@{}}
    \toprule
    Once: SAM~3 + SAM~3D on frame 0 & 5.5\,s \\
    Per frame: SAM~2 + rendering & 35\,ms \\
    Per frame: VAE encoding & 79\,ms \\
    Per frame: DiT (4 steps per chunk) & 84\,ms \\
    Per frame: VAE decoding & 79\,ms \\
    \midrule
    1 GPU: FPS / first chunk & 3.6 / 2.5\,s \\
    4 GPUs: FPS / first chunk & 11.2 / 2.4\,s \\
    \bottomrule
  \end{tabular}
\end{table}

\subsection{Runtime}
\label{sec:supp_runtime}
\Cref{tab:supp_runtime} lists the cost of condition construction and video generation on one NVIDIA RTX PRO 6000 Blackwell Max-Q GPU.
For a source video, the bidirectional model needs the full conditions before generation: anchor grounding and tracking, reconstruction, camera estimation with ViPE, alignment and rendering, about 3 minutes per clip on an RTX 6000 Ada.
In the streaming setting, the anchor is placed once from the first frame with the pose predicted by SAM~3D, and only SAM~2 tracking and the rendering run per frame.
The FPS of the autoregressive models in \cref{tab:co3d} is this streaming rate, including the per-frame conditions, while the FPS of the bidirectional models counts generation only.
Running the condition stage, the VAE encoder, the DiT and the VAE decoder of {\ourmethodar} on four GPUs as a pipeline raises the throughput to 11.2 FPS with identical outputs; the first chunk still appears after 2.4\,s, since it passes through every stage.

\end{document}